\documentclass[10pt,journal,compsoc]{IEEEtran}
\ifCLASSOPTIONcompsoc
  \usepackage[nocompress]{cite}
\else
  \usepackage{cite}
\fi

\ifCLASSINFOpdf
   \usepackage[pdftex]{graphicx}
\else
\fi

\usepackage{xcolor}
\usepackage{helvet}
\usepackage{courier}
\usepackage{graphicx}
\usepackage{subfigure}
\usepackage{balance}
\usepackage{amsmath,amsthm,amssymb}
\usepackage{textcomp,booktabs}
\usepackage{footmisc}
\usepackage[normalem]{ulem}
\usepackage{multirow}
\usepackage{setspace}
\usepackage{bigstrut}
\usepackage{array}
\usepackage{xcolor}
\usepackage{amsfonts}
\usepackage{bm}
\usepackage{url}

\newcommand{\x}{\boldsymbol{x}}

\newcommand{\w}{\boldsymbol{w}}

\newcommand{\X}{\mathbf{X}}

\newcommand{\y}{\boldsymbol{y}}

\newcommand{\p}{\boldsymbol{p}}

\newcommand{\Dm}{\mathcal{D}}

\IEEEpubid{
\fbox{
\parbox{0.98\textwidth}{
\copyright 2020 IEEE. Personal use of this material is permitted. Permission from IEEE must be obtained for all other uses, in any current or future media, including\hfill
reprinting/republishing this material for advertising or promotional purposes, creating new collective works, for resale or redistribution to servers or lists, or reuse of\hfill
any copyrighted component of this work in other works. DOI: 10.1109/TPAMI.2020.2964173\vspace{-0.5mm}
}%
}}
\begin{document}
% \begin{CJK}{UTF8}{song}

%
\title{Deep Residual Correction Network for \\Partial Domain Adaptation}

\author{Shuang Li, Chi Harold Liu,~\IEEEmembership{Senior Member,~IEEE,} Qiuxia~Lin, Qi~Wen, Limin~Su, \\Gao~Huang and Zhengming~Ding
\IEEEcompsocitemizethanks{
\IEEEcompsocthanksitem S. Li, C. H. Liu, Q. Lin, Q. Wen and L. Su are with the School of Computer Science and Technology, Beijing Institute of Technology, Beijing, China. Corresponding author: C. H. Liu. Email: \{shuangli,chiliu,linqiuxia,qwen,sulimin\}@bit.edu.cn\protect\\
\IEEEcompsocthanksitem G. Huang is with Department of Automation, Tsinghua University, Beijing, China. Email: gaohuang@tsinghua.edu.cn \protect\\
\IEEEcompsocthanksitem Z. Ding is with Department of Computer, Information and Technology, Indiana University-Purdue University Indianapolis, USA. Email: zd2@iu.edu \protect\\
}
}
%E-mail: see http://www.michaelshell.org/contact.html

\markboth{IEEE TRANSACTIONS ON PATTERN ANALYSIS AND MACHINE INTELLIGENCE,~Vol.~XX, No.~XX, XXX~201x}%
{Li \MakeLowercase{\textit{et al.}}: Deep Residual Correction Network for Partial Domain Adaptation}

\IEEEtitleabstractindextext{%
\begin{abstract}
Deep domain adaptation methods have achieved appealing performance by learning transferable representations from a well-labeled source domain to a different but related unlabeled target domain. Most existing works assume source and target data share the identical label space, which is often difficult to be satisfied in many real-world applications. With the emergence of big data, there is a more practical scenario called \textit{partial domain adaptation}, where we are always accessible to a more large-scale source domain while working on a relative small-scale target domain. In this case, the conventional domain adaptation assumption should be relaxed, and the target label space tends to be a subset of the source label space. Intuitively, reinforcing the positive effects of the most relevant source subclasses and reducing the negative impacts of irrelevant source subclasses are of vital importance to address partial domain adaptation challenge. This paper proposes an efficiently-implemented \textit{Deep Residual Correction Network} (DRCN) by plugging one residual block into the source network along with the task-specific feature layer, which effectively enhances the adaptation from source to target and explicitly weakens the influence from the irrelevant source classes. Specifically, the plugged residual block, which consists of several fully-connected layers, could deepen basic network and boost its feature representation capability correspondingly. Moreover, we design a weighted class-wise domain alignment loss to couple two domains by matching the feature distributions of shared classes between source and target. Comprehensive experiments on partial, traditional and fine-grained cross-domain visual recognition demonstrate that DRCN is superior to the competitive deep domain adaptation approaches.
%Promisingly, we obtain around 10\% improvements over comparisons on some partial domain adaptation tasks.
\end{abstract}

\begin{IEEEkeywords}
Deep Transfer Leaning, Partial Domain adaptation, Maximum Mean Discrepancy, Fine-grained Visual Recognition.
\end{IEEEkeywords}}

\maketitle

\IEEEdisplaynontitleabstractindextext
\IEEEpeerreviewmaketitle

\IEEEraisesectionheading{\section{Introduction}\label{sec:introduction}}

\IEEEPARstart{N}{umerous} recent advances have dramatically improved the performance of deep neural networks (DNNs) at several diverse learning tasks such as network designing\cite{imagenet,IGC,ResNext,IGC2}, computer vision \cite{DL,fine-grained1,fine-grained2,DSN,Large-Scale}, and natural language processing \cite{Attention,S2S,CS2S}, etc. Nevertheless, most state-of-the-art models with appealing performance mainly rely on the massive amount of annotated data, however, it is always time-consuming and expensive to obtain sufficient amount of labeled training data \cite{survey,ARCPIP,DA-survey}. At the same time, traditional machine learning methods often have one common assumption that the training and test data are drawn from the same or similar probability distribution, which cannot always be satisfied in the real-world applications. Therefore, there is a strong motivation to leverage the useful knowledge of a related labeled \textit{source domain} to help design versatile models for our interested unlabeled \textit{target domain} following different feature distributions. To achieve this, domain adaptation \cite{survey,TCA} is a promising strategy to address the domain shift issue, and impressive progress has been made in a wide range of scenarios \cite{JAN,DA_bp,co-training,MSDA,DAN,fine-grained-TL,TL-MV,MTL-head}.

The mainstream methods of domain adaptation are either instance reweighting based methods \cite{KMM,PRDA} by assigning different weights to labeled source samples, or domain-invariant feature learning based methods \cite{JDA,JGSA,DICD} by mitigating the distributions discrepancy across two domains. On the contrary, recent advance in deep learning reveals that deep architectures can extract more transferable and domain-invariant representations when dealing with domain adaptation problems \cite{DeCAF,DAN,how-transfer-feature}. Moreover, deep learning based methods have achieved superior performance over most shallow domain adaptation approaches.

However, most of the current deep domain adaptation methods still assume the source and target domains share one identical label space. In other words, data in both domains belong to exactly the same classes. Then, by aligning statistic moments \cite{DDC,LDC,DAN,RTN,JAN} or leveraging adversarial techniques \cite{DA_bp,MADA,SDT,ADDA}, they could mitigate the marginal distribution differences across two domains to learn transferable representations. For feasible adaptation, the label spaces of both domains are required to be identical, otherwise the data in the irrelevant source subclasses will mislead the target data alignment and cause \textit{negative transfer} when recognizing target classes.

\begin{figure}[tb]
\centering
\includegraphics[width=0.49\textwidth]{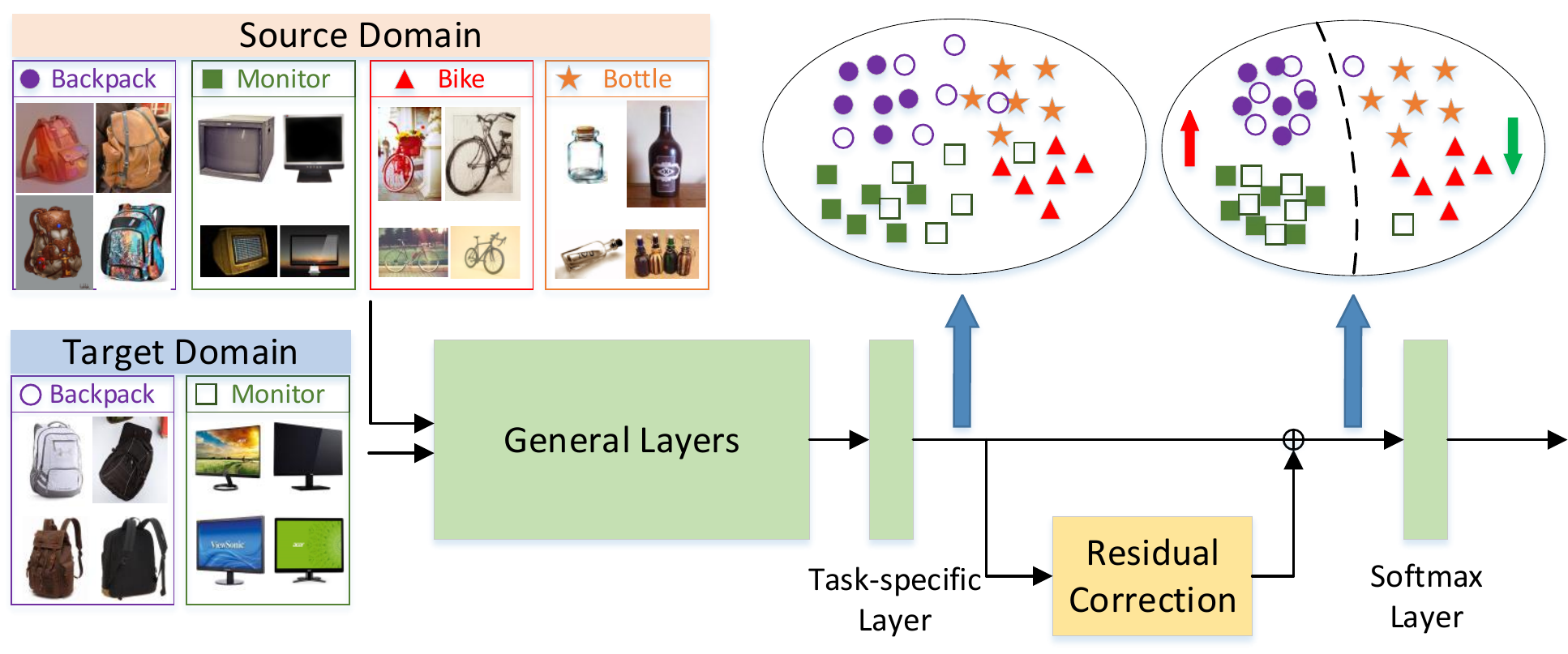}\vspace{-5mm}
\caption{Illustration of the proposed \textit{Deep Residual Correction Network} (DRCN) approach that aims to address partial domain adaptation problem, where the label space of the interested small-scale target domain is a subset to that of a large-scale source domain. By utilizing the weighted class-wise matching, DRCN can effectively select out the irrelevant source classes and enhance the importance scores of the most relevant source classes. In addition, the plugged residual block could adaptively align source and target domains, and boost its feature representation capability, which results in better target classification performance.}\vspace{-3mm}
\label{Fig1_motivation}
\end{figure}

Most recently, a more challenging and practical scenario, referred to as \textit{partial domain adaptation}, attracts a lot of attentions \cite{PADA,SAN,WAN-for-partial}, where a large-scale source domain is diverse enough to subsume all classes in a small-scale target domain of interest. Furthermore, target data are unlabeled and we have no idea about the size of target classes or the corresponding categories. Intuitively, to address partial domain adaptation problems, we cannot simply align the whole source and target domains, since the irrelevant source subclasses will be mixed with target data together, resulting in the degradation of target classification performance. Thus, moving out the irrelevant source classes and enhancing the effects of the most similar source classes with target domain are crucial for effective knowledge transfer. To achieve this purpose, \cite{SAN,PADA} propose to maximally match both domains' distributions in the shared label space and diminish the negative impact of irrelevant source classes. \cite{WAN-for-partial} designs a novel adversarial domain classifier to identify the importance weight of each source data automatically. However, these methods all utilize adversarial techniques to align source and target, which are difficult to optimize comparing with MMD loss based domain adaptation methods.

In this paper, we propose an efficiently-implemented \textit{Deep Residual Correction Network} (DRCN) to address partial domain adaptation problem by identifying desirable classes and strengthening feature transfer. On the one hand, we present weighted class-wise domain alignment to automatically uncover the most relevant source classes to target data according to the target output probability distribution, and assign larger weights to maximally facilitate the class-wise alignment between these shared classes across domains.
Besides, since the deep features in standard CNN architectures transit from general to specific along the network, the cross-domain discrepancy between source and target will increase, which results in the transferability decreasing of task-specific layers \cite{how-transfer-feature}. Therefore, we expect to enhance the domain shift mitigation between the two domains through the designed residual correction block. As illustrated in Fig. \ref{Fig1_motivation}, we add one residual correction block along with the task-specific feature layer in a general network. Contributed by the class-wise alignment and enhancement of residual block, DRCN can maximally reduce the disparity of target data with those source data whose classes are likely to appear in the target label space.

On the other hand, when classifier prediction is quite poor in hard tasks, we believe domain-wise knowledge containing general information is also necessary in partial domain adaptation scenarios.
%thereby greatly transferring the general feature-level and task-level knowledge from source to target.
To achieve this goal, DRCN aligns the joint distributions of last two domain-specific layers across domains by minimizing a joint maximum mean discrepancy (JMMD) metric \cite{JAN}, thereby greatly transferring the general feature-level and task-level knowledge from source to target. To sum up, we have four-fold contributions as follows:\vspace{-1mm}
\begin{itemize}
%  \item We propose an easily implemented but effective \textit{Deep Residual Correction Network} (DRCN) to address the partial domain adaptation problem, which faces more realistic and challenging scenarios. Specifically, we propose a weighted class-wise alignment loss which can identify the most relevant source subclasses to target data based on the target output probability distribution. Larger weights will be assigned to these classes to benefit the accurate class-wise matching across domains a lot. To enhance knowledge transfer, we design an effective residual block to explicitly measure and learn the difference across domains by exploiting the structure.
  \item We propose an easily implemented but effective \textit{Deep Residual Correction Network} (DRCN) to address the partial domain adaptation problem, which faces more realistic and challenging scenarios. DRCN only requires to plug one residual block into a unified network to mitigate the cross-domain distributions discrepancy. The added residual block can explicitly capture the feature difference between source and target, which intrinsically addresses the inherent problem in partial domain adaptation scenarios.
  \item DRCN introduces a weighted class-wise alignment loss which can automatically identify the most relevant source subclasses to target data based on the target output probability distribution. Larger weights will be assigned to these classes to benefit the accurate class-wise matching across domains a lot.
  %could not only transfer general feature-level and task-level knowledge from source to target, but also automatically identify the most relevant source subclasses to target data based on the target output probability distribution. Larger weights will be assigned to these classes to benefit the accurate class-wise matching across domains a lot.
  \item Moreover, DRCN properly explores domain-wise knowledge to adapt global source information to target, which is a different attempt from the existing partial domain adaptation methods. When a large cross-domain discrepancy exists, the leveraging of domain-wise knowledge is crucial for effectively solving partial domain adaptation problems.
  \item Furthermore, DRCN can be easily extended to deal with traditional and fine-grained cross-domain visual recognition, in which source and target domains share one identical label space or even fined-grained visual data. This manifests the universality of the proposed DRCN. Extensive experiments on partial, traditional and fine-grained cross-domain visual recognition demonstrate DRCN outperforms other competitive comparisons with a large margin.
\end{itemize}

\section{Related Work}\label{sec:relatedwork}

Recent progresses in domain adaptation \cite{survey} are able to address the issue that training and test data follow different feature spaces and data distributions. This helps mitigate the burden of manual labeling by exploring the external source knowledge, which promotes impressive research efforts in lots of applications \cite{TCA,NLP-S,CORAL,DA_bp,PADA,SAN}.

\subsection{Domain Adaptation}

Extensive prior works on domain adaptation usually attempt to minimize the domain discrepancy through instance reweighting or domain-invariant feature learning \cite{survey}. Instance reweighting based methods are encouraged to align source and target distribution by reweighting source samples\cite{KMM,co-training}. However, these methods exist limitation when training and test data follow different conditional distributions. On the other hand, feature learning based methods aim to derive domain-invariant features or latent subspaces, by reducing the distribution differences across domains. For example, Scatter Component Analysis (SCA, \cite{SCA}) seeks to trade between maximizing the separability of classes, minimizing the mismatch between domains, and maximizing the separability of data. TCA \cite{TCA} considers learning some transfer components in RKHS to minimize the distance of source and target marginal distributions using Maximum Mean Discrepancy (MMD) metric. Benefiting from leveraging target pseudo labels, JDA \cite{JDA} attempts to match both marginal and conditional distributions of two domains by requiring their total means and class means to be close to each other. However, these methods all can only learn shallow features for both domains, which are not effective enough comparing with deep learning based domain adaptation methods.
%However, these methods all fail to jointly optimize the domain-invariant features and target labels refinement. Therefore, these two procedures learned separately cannot benefit each other in an effective way and can only learn shallow features for both domains.
% feature extraction approach.like feature  which requires expensive data-labeling efforts.

\subsection{Deep Domain Adaptation}

Deep learning can learn abstract representations that disentangle different explanatory factors of variations behind data \cite{review-representation} and manifest invariant factors underlying different populations that transfer well across similar tasks \cite{how-transfer-feature}. However, some recent findings reveal that deep networks can only reduce, but not eliminate, the cross-domain discrepancy \cite{how-transfer-feature,DDC}.
%, nor removing the cross-domain discrepancy

Hence, there are several recent attempts in bridging deep learning with domain adaptation, which are generally achieved by adding adaptation layers through which the means of distributions are matched \cite{DDC,RTN} or adding a subnetwork as a domain discriminator which would be confused by learned deep features from two domains \cite{SDT,DA_bp}. Some previous works on deep domain adaptation usually attempt to align two domains by utilizing MMD metric. To name a few, Tzeng \emph{et al.} in \cite{DDC} utilize an adaptation layer along with a domain confusion loss based on MMD to automatically learn the domain-invariant representations jointly trained with a classifier. In \cite{DAN}, Long \emph{et al.} leverage multi-layer adaptation via multi-kernel MMD, which is able to align different domains in the task-specific layers. Being primarily motivated by ResNet in \cite{resnet}, \cite{RTN} proposes Residual Transfer Networks (RTN) to bridge the source classifier $f_s(x)$ and target classifier $f_t(x)$ with the residual layers, such that the classifier mismatch across domains can be explicitly modeled by the residual functions $\Delta F(x)$ in a deep learning architecture. Different from RTN, our proposed residual block aims to explicitly learn and mitigate the feature difference, instead of the classifier difference, between source and target, which can intrinsically address the inherent problem in domain adaptation scenarios. In addition, we find that it is more effective to mitigate the domain feature discrepancy rather than classifier difference. Furthermore, Joint Adaptation Networks (JAN) \cite{JAN} develops a joint maximum mean discrepancy (JMMD) criterion which can be calculated by back-propagation in linear-time.
%However, all these methods mentioned above mainly correct the shifts in marginal distributions, assuming conditional distributions remain unchanged, except that JAN considers the joint probability distribution.

Another line of methods are based on an adversarial loss which introduces a novel domain classifier to promote domain confusion, where the data would be indiscriminative with respect to domain labels. Recent works \cite{ADDA,SDT,DA_bp}, combining adversarial learning with domain adaptation, have shown significantly improved performance. Specifically, \cite{ADDA} outlines a novel generalized adversarial adaptation framework, which subsumes several deep adversarial transfer learning methods as special cases. Based on this framework, \cite{ADDA} also develops an Adversarial Discriminative Domain Adaptation (ADDA) approach.
%Specifically, these adversarial adaptation methods are based on the idea of Generative Adversarial Network (GAN, \cite{GAN}). GAN is a generative deep model that trains two networks with competing goals.
%Following this line, Salimans \emph{et al.} in \cite{train-GAN} introduce several techniques to stabilize GAN training, and an evaluation metric providing a basis for comparing the quality of GANs. % Liu \emph{et al.} in \cite{CoGAN} presented a coupled generative adversarial network (CoGAN) that can learn a joint distribution without corresponding images in different domains, by enforcing a weight-sharing constraint to the framework. In contrast, InfoGAN \cite{infoGAN} is an information-theoretic extension to GAN that is able to learn disentangled representations and maximize the mutual information between a small subset of noise variables and observations for GAN.
By exploring the adversarial loss in aligning two domains in terms of feature level or image level, there are many research efforts done recently \cite{SDT,DA_bp,DSN-TL,pixel-level-GAN}. The general idea behind is to generate domain-invariant features across two domains by confusing the discriminator. Usually minimax strategy is adopted to optimize two players in adversarial learning based methods, thus it is usually hard to achieve stable solutions compared with MMD-based methods.

However, these methods all assume the source and target label spaces are the same, which is often too strict to be satisfied in real-world applications. The proposed DRCN has relaxed this common assumption, and aims to address a more challenging partial domain adaptation problem.

% namely, this discriminator cannot determine which domain the data come from. In this manner, these two domains are considered to be drawn from the same distribution

% GAN
% In related work, adversarial learning has been explored for generative tasks. The Generative Adversarial Network (GAN) method\cite{GAN} is a generative deep model that pits two networks against one another:

% Recent domain adaptation methods learn deep neural transformations that map both domains into a common feature space. Recent work has focused on transferring deep neural network representations from a labeled source datasets to a target domain where labeled data is sparse or non-existent. In the case of unlabeled target domains (the focus of this paper) the main strategy has been to guide feature learning by minimizing the difference between the source and target feature distributions

% optimizing the representation to minimize some measure of domain shift such as maximum mean discrepancy or correlation distances. An alternative is to reconstruct the target domain from the source representation
% some recent work bridges deep learning and domain adaptation, which extends deep convolutional networks to domain adaptation layers through which the mean embeddings of distributions are matched.

\subsection{Partial Domain Adaptation}

In reality, we are much easier to obtain a large-scale source dataset, while working on a small-scale unlabeled target dataset, which requires us to transfer partial relevant knowledge from the source to target. In this way, previous domain adaptation methods \cite{DAN,ADDA}, typically assuming that source and target domains have identical label space, are prone to \textit{negative transfer} for the partial transfer problems.
%Most recently, some significant advances proposed the novel and realistic application scenario, which is related to partial transfer learning. Along this line,

To effectively address partial domain adaptation problems, Cao \emph{et al.} in \cite{SAN} presents a Selective Adversarial Network (SAN), which can promote positive transfer of relevant data and alleviate negative transfer of irrelevant data simultaneously, and maximally match the data distributions in the shared class space. Partial Adversarial Domain Adaptation (PADA) \cite{PADA} alleviates the negative transfer by down-weighing the data of outlier source classes. Zhang \emph{et al.} propose a novel adversarial nets-based partial domain adaptation method to identify the source samples that are potentially from the outlier classes, while reducing the domain shift of shared classes between domains \cite{WAN-for-partial}. These methods all rely on various adversarial strategies and networks to learn the target model.
%These methods all consider mitigating irrelevant source class influence on feature transfer across domains by weighting.

Different from them, our DRCN only plugs one residual correction block into source network to explicitly mitigate the domain discrepancy between domains. By leveraging the target output probability distribution, DRCN could effectively identify the most relevant source subclasses and maximally conduct feature alignment through the proposed weighted class-wise matching scheme, which is clearly different from other partial domain adaptation methods.

\section{The Proposed Algorithm: DRCN}\label{sec:method}
This section introduces our proposed Deep Residual Correction Network (DRCN) in detail. The background and preliminary knowledge will be presented first.

\subsection{Preliminary}

Partial domain adaptation is a novel but practical transfer learning paradigm \cite{SAN,PADA}, which manages to transfer relevant knowledge from a large-scale source domain to a small-scale target domain. Similar to traditional domain adaptation terminologies, in partial domain adaptation, labeled source domain $\Dm_s$ and unlabeled target domain $\Dm_t$ are provided, where $\Dm_s=\{({\x_s}_i,{y_s}_i)\}_{i=1}^{n_s}=\{\X_s,\y_s\}$, $\mathcal{D}_t=\{{\x_t}_j\}_{j=1}^{n_t}=\{\X_t\}$. ${y_s}_i$ is the corresponding label of ${\x_s}_i$. $n_s$, $n_t$ are the numbers of source and target samples.

Suppose $\mathcal{X}_s, \mathcal{X}_t$ and $\mathcal{Y}_s, \mathcal{Y}_t$ are the feature and label spaces of source and target domains, respectively. For partial domain adaptation, $\mathcal{X}_s=\mathcal{X}_t$, while the target label space $\mathcal{Y}_t$ is a subset of the source label space $\mathcal{Y}_s$, i.e., $\mathcal{Y}_t \subseteq \mathcal{Y}_s$. Here we denote $C_s, C_t$ as the class numbers of source and target domain, and $C_s>C_t$. By contrast, $C_s$ equals to $C_t$ in the traditional domain adaptation scenario. In addition, due to the domain shift, the source and target feature distributions $P_s(\x)\neq P_t(\x)$. The goal is to maximally mitigate the partial distribution discrepancy across domains, and transfer the most relevant source discriminative knowledge to target domain effectively.

\subsection{Motivation}

Since partial transfer learning assumes target label space subsumes to source label space, and the distributions of both domains with the shared classes are also different. It is of vital importance to figure out \textit{what knowledge should be transferred from a large-scale source to a small-scale target}, and \textit{how to capture the relevant knowledge effectively}?

Intuitively, when the categories of target data are only a subset of source data, we should avoid only matching the distributions of the whole source and target domains, since the irrelevant source classes will have negative effect on the domain alignment. As a result, the most relevant source classes to target domain should be effectively identified, and precisely align their conditional distributions between two domains. Nevertheless, we also believe the generic information learned by pre-trained network on massive source data is valuable and crucial for partial domain adaptation.
Additionally, because task-specific layer contains feature-level knowledge while classifier layer contains rich multimodal structure which is task-level knowledge, we therefore want to simultaneously align these layers to greatly transfer general feature-level and task-level knowledge.
To achieve the aforementioned goals, we propose to align the joint distributions between input features and output labels across domains, as well as the class-wise distributions between source and target in the shared class space.

Recent works of deep domain adaptation have revealed that many consecutive layers of non-linear transformations within the trained source network will amplify the feature distributions difference across domains \cite{how-transfer-feature,DAN}. Thus, an effective and direct way to compensate the domain shift is to correct the feature discrepancy of source and target right along the task-specific layer. Inspired by the well-known residual network \cite{resnet}, we aim to plug one residual correction block, consisting of several layers, into the trained source network to explicitly learn the feature difference between source and target domains. In addition, the added residual block could improve the generalization ability of the infrastructural network by deepening it. This small modification of source network has been proven to be very effective in our experiments.

\subsection{Deep Residual Correction Network}
% In this paper, we will utilize maximum mean discrepancy (MMD)\cite{MMD} metric to facilitate the feature adaptation across domains. Thus, we first introduce MMD, and the details of DRCN will be described later.
We first revisit maximum mean discrepancy (MMD) \cite{MMD} metric to facilitate the feature adaptation across domains, and then present the details of our proposed DRCN.

\subsubsection{Maximum Mean Discrepancy Revisit}

MMD statistical test is commonly leveraged to quantitatively measure the similarity of source and target probability distributions $P_s(\x)$ and $P_t(\x)$ \cite{DAN,RTN,JAN}. If we denote $\mathcal{F}$ is a universal class of functions, MMD can be formally represented as:
\begin{align}\label{eq1:MMD-denifination}
\mathcal{D}_{\mathrm{MMD}} (P_s,P_t)= \sup_{f\in\mathcal{F}}\left(\mathbb{E}_{\x\sim P_s}[f(\x)]-\mathbb{E}_{\x\sim P_t}[f(\x)]\right)^2,
\end{align}
where $f$ is a function from $\mathcal{F}$. It has been theoretically proven that $P_s(\x)$ and $P_t(\x)$ are identical if and only if $\mathcal{D}_{\mathrm{MMD}} (P_s,P_t)=0$ \cite{MMD}.

If $\mathcal{F}$ is a universal RKHS with kernel function $\kappa(\cdot,\cdot)$, (\ref{eq1:MMD-denifination}) can be represented as
% \begin{small}
\begin{align}
\label{eq2:MMD-closed}
&\mathcal{D}_{\mathrm{MMD}} (P_s,P_t)=\mathbb{E}
_{\x_s,\x_s'\sim P_s}[\kappa(\x_s,\x_s')]\nonumber\\
&-2\mathbb{E}_{\x_s\sim P_s,\x_t\sim P_t}[\kappa(\x_s,\x_t)]+\mathbb{E}_{\x_t,\x_t'\sim P_t}[\kappa(\x_t,\x_t')].
\end{align}
% \end{small}
In (\ref{eq2:MMD-closed}), $\x_s,\x_s'$ and $\x_t,\x_t'$ are samples drawn from distributions $P_s(\x)$ and $P_t(\x)$ respectively. When we have finite source and target samples, the MMD distance of both domains can be estimated as
%% \begin{small}
%\begin{align}
%\label{eq3:MMD-estimated}
%\widehat{\mathcal{D}}_{\mathrm{MMD}} (P_s,P_t)=&\frac{1}{n_s^2}\sum_{i,j=1}^{n_s} \kappa({\x_s}_i,{\x_s}_j)\nonumber\\
%&-\frac{2}{n_sn_t}\sum_{i=1}^{n_s}\sum_{j=1}^{n_t}\kappa({\x_s}_i,{\x_t}_j)\nonumber\\
%&+\frac{1}{n_t^2}\sum_{i,j=1}^{n_t}\kappa({\x_t}_i,{\x_t}_j)
%\end{align}
%% \end{small}
\begin{align}
\label{eq3:MMD-estimated}
&\widehat{\mathcal{D}}_{\mathrm{MMD}} (P_s,P_t)=\frac{1}{n_s^2}\sum_{i,j=1}^{n_s}\kappa({\x_s}_i,{\x_s}_j)\nonumber\\
&-\frac{2}{n_sn_t}\sum_{i=1}^{n_s}\sum_{j=1}^{n_t}\kappa({\x_s}_i,{\x_t}_j)
+\frac{1}{n_t^2}\sum_{i,j=1}^{n_t}\kappa({\x_t}_i,{\x_t}_j)
\end{align}

MMD has been successfully applied in massive areas of deep learning, ranging from generative adversarial models \cite{GAN,GAN-MMD} to image transformation \cite{DMT}. The general idea behind using MMD in this paper is to explicitly depict the distribution difference between source and target domains. By minimizing the improved MMD metric \cite{DAN,JAN}, we could achieve effective feature and task adaptation.

\subsubsection{Residual Correction Block}

\begin{figure}[tb]
\centering
\includegraphics[width=0.5\textwidth]{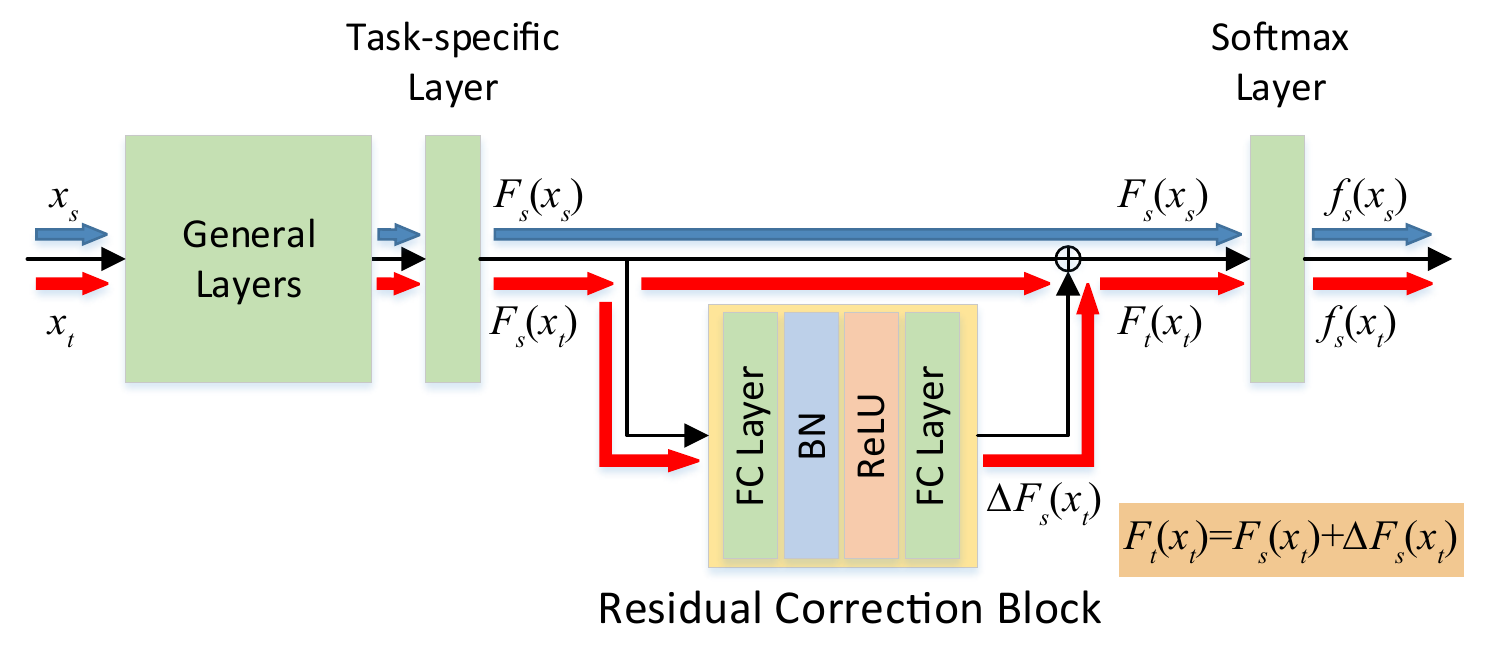}\vspace{-3mm}
\caption{The architecture and data flow of the added residual correction block in a pre-trained source network, where a FC Layer is a fully-connected layer with appropriate dimensions, BN represents the Batch Normalization layer \cite{BN}, and ReLU is the non-linear transformation. The blue and red arrows denote the data flow of source and target data, respectively. DRCN expects the added residual correction block to learn the difference $\Delta F_s(\x_t)$ between feature representations $F_s(\x_s)$ and $F_s(\x_t)$, such that $F_s(\x_s)$ and $F_t(\x_t)$ can be more similar.}
\label{Fig2_residualblock}\vspace{-3mm}
\end{figure}

Deep architectures can learn more abstract, transferable and desirable features automatically \cite{how-transfer-feature,DAN}. However, the learned deep features can only reduce, but not remove, the substantial cross-domain discrepancy. Especially, the consecutive non-linear transformation in a deep network may amplify the difference between domains.

One intuitive way to mitigate the cross-domain discrepancy is to correct it right after the task-specific layer of a source network, since recent studies have revealed that the feature transferability drops from general feature extraction layers to higher task-specific layers under the domain discrepancy \cite{how-transfer-feature}. Meanwhile, the features learned in the higher task-specific layers are more crucial for the final classification. Therefore, as shown in Fig. \ref{Fig2_residualblock}, after the task-specific layer in a pre-trained source network, we plug one residual correction block to explicitly learn the cross-domain discrepancy.
%Instead of just deploying discrepancy loss, we explicitly measure and learn the difference by exploiting the structure.
Most works align features in task-specific layer just by deploying discrepancy losses, whereas we find that exploiting the structure modification can further explicitly measure and learn the difference.
Our designed residual correction block will not only keep the feature extraction and discriminative classification abilities of the original source network, but also benefit the deep adaptation from source to target of interest.

As illustrated in Fig. \ref{Fig2_residualblock}, we denote the task-specific features of source data $\x_s$ and target data $\x_t$ as $F_s(\x_s)$ and $F_s(\x_t)$, respectively. $\x_s$ only goes through the original source network. Besides the source network, $\x_t$ also passes the added residual correction block, which aims to learn the domain discrepancy $\Delta F_s(\x_t)$. Then, we expect that the modified target representations $F_t(\x_t)=F_s(\x_t)+\Delta F_s(\x_t)$ will be much more similar to source representations $F_s(\x_s)$ after the adaptation. Because we only let target sample pass through residual correction block and align it with source sample, such a weakly-shared structure allows target domain to model its difference with source domain effectively and properly. Besides, the added residual block deepens the general network, which could improve its feature representation capability correspondingly.

To this end, in DRCN, we first require the empirical error of the source network classifier $f_s$ on labeled source data $\{({\x_s}_i,{y_s}_i)\}_{i=1}^{n_s}$ to be minimized:
\begin{align}
\label{eq4:source-loss}
\min_{f_s}~~~~\mathcal{L}_s=\frac{1}{n_s}\sum_{i=1}^{n_s}\mathcal{L}(f_s({\x_s}_i),{\y_s}_i),
\end{align}
where ${\y_s}_i$ is the one-hot label for ${\x_s}_i$ and $\mathcal{L}(\cdot,\cdot)$ is the cross-entropy loss function defined as follows:
\begin{align}
\label{eq5:source-cross-entropy}
\mathcal{L}(f_s({\x_s}_i),{\y_s}_i)=-\sum_{k=1}^{C_s}\mathbf{1}_{[k={y_s}_i]}\log {f_s}^{(k)}({\x_s}_i),
\end{align}
where ${f_s}^{(k)}({\x_s}_i)$ is the probability of ${\x_s}_i$ predicted to the $k$-th class by $f_s$.

Similar to \cite{DAN,WAN-for-partial,JAN}, we adapt the pre-trained source network from source data to a small-scale target domain.
For partial domain adaptation, we also need to select out the irrelevant source subclasses, and assign larger importance scores to these most relevant source data. To achieve this goal, DRCN could automatically calculate the importance of each source class according to the target output probability distribution, and explicitly conduct weighted class-wise distribution matching to transfer the most related and useful knowledge from source to target.
%To maximally preserve the characteristics and knowledge of a large-scale source domain, we opt to align features and classifiers of both domains in a general way first, which could transfer the basic knowledge from source to target.
%Simultaneously, we will identify the source classes which are more potential to appear in the target label space and conduct weighted class-wise matching between them. This strategy could benefit improving the positive transfer from source to target and transferring more accurate discriminative information to target effectively.

\subsubsection{Weighted Class-wise Alignment}

Different from the traditional domain adaptation settings, where the label spaces of both domains are identical, partial domain adaptation assumes target domain only contains a subset of classes in source domain. Therefore, identifying the most relevant source subclasses to target, and conducting class-wise feature distributions alignment of shared classes across domains are crucial to derive perfect partial domain adaptation models.

In this paper, we leverage the target output probability distribution, referred to as target ``soft label'' \cite{SDT} in each category, predicted by the source classifier, to identify the most possible classes that target data contain. Then, we can select the corresponding source subclasses to conduct class-wise matching. To be specific, for each target data ${\x_t}_j$, its prediction by the source classifier $f_s$ is denoted as ${\p_t}_j$. ${\p_t}_j$ is a label vector, whose $k$-th element ${\p_t}_j^{(k)}$ represents the probability of assigning ${\x_t}_j$ to class $k$ in $C_s$ classes. If we average all target predictions from $f_s$, we can obtain the target output probability distribution over $C_s$ classes, which could effectively uncover the proportion of each target class in a larger source label space.

The average of all the target predictions could be computed as:
\begin{align}
\label{eq8:class-weight}
\w=\frac{1}{n_t}\sum_{j=1}^{n_t}{\p_t}_j,
\end{align}
which could be used to assign different importance scores to different classes. From (\ref{eq8:class-weight}), we observe that if the $k$th element of $\w$, i.e. $\w^{(k)}$ is sufficiently small, it means all the target data are predicted to class $k$ with small probability, and the class $k$ of source data should be irrelevant to the small-scale target domain with large probability. Thus, we should assign a small weight to class $k$ when we conduct class-wise feature alignment, and vice versa. In practice, we normalize the weight $\w$ by dividing its maximum element as \cite{PADA}, i.e.
\begin{align}
\label{eq9:class-weight-normalize}
\w\leftarrow\frac{1}{\mathrm{max}(\w)}\w.
\end{align}

After identifying the most relevant source subclasses automatically, we conduct weighted class-wise distribution matching to improve the positive effects of relevant source classes and reduce the negative impacts of irrelevant source classes, which could benefit transferring valuable knowledge from source to target a lot.

To be specific, we leverage multiple kernel variant of MMD (MK-MMD) \cite{MMD,DAN} to effectively reduce the discrepancy of each class across domains. The squared formulation of MK-MMD distance for source and target data representations of class $k$ can be estimated as:
\begin{small}
\begin{align}
\label{eq10:MKMMD}
&\widehat{\mathcal{D}}^{(k)}_{\mathrm{MK-MMD}}=\nonumber\\
&\bigg\Vert \frac{1}{n_s^{(k)}}\sum_{{\x_s}_i\in\Dm_s^{(k)}}\phi \big(F_s({\x_s}_i)\big)
-\frac{1}{n_t^{(k)}}\sum_{j=1}^{n_t}{\p_t}_j^{(k)}\phi\big(F_t({\x_t}_j)\big) \bigg\Vert^2_{\mathcal{H}_{\kappa}},
\end{align}
\end{small}
where $n_t^{(k)}=\sum_{j=1}^{n_t}{\p_t}_j^{(k)}$, $\mathcal{D}_s^{(k)}$ denotes all the source samples with their true labels being $k$, $\mathcal{H}_{\kappa}$ is the RKHS with a characteristic kernel $\kappa$. $\phi$ is the corresponding feature map.

For partial domain adaptation problems, we should apply the class weight, computed in (\ref{eq9:class-weight-normalize}) to improve the contributions of relevant classes adaptation between two domains, and down-weigh the contributions of irrelevant classes alignment when learning the final model. Therefore, the loss term of weighted class-wise distribution matching in DRCN can be formulated as
\begin{align}
\label{eq11:weighted-class-mmd}
\mathcal{L}_{class}=\sum_{k=1}^{C_s}\w^{(k)}\cdot\widehat{\mathcal{D}}^{(k)}_{\mathrm{MK-MMD}}.
\end{align}

By minimizing (\ref{eq11:weighted-class-mmd}), DRCN could put more emphasis on the shared classes with larger weights, and transfer the valid feature and task knowledge from source to target without distraction of irrelevant source classes. Different from other partial domain adaptation \cite{PADA,SAN,WAN-for-partial}, we propose to explicitly match class-wise distributions across domains in the shared label space by minimizing a novel probabilistic class-wise multiple kernel MMD, which considers prior target category distributions to reduce the conditional distribution divergence in an effective way.

However, we find that general information is also crucial for partial domain adaptation, as class-wise alignment might not work well when model predication is very poor. Therefore, to maximally preserve the characteristics and general knowledge of a large-scale source domain, we opt to align features and classifiers of both domains in a general way, which facilitates transferring the basic knowledge from source to target.

\subsubsection{General Feature Adaptation}
It is noteworthy that, in contrast to other partial domain adaptation methods \cite{SAN,PADA,WAN-for-partial}, DRCN also aligns the joint distributions of both domains to some extent. \cite{SAN,PADA,WAN-for-partial} claim that the irrelevant source subclasses will degrade the adaptation performance, and thus, they do not intend to align the whole source and target domains.

Nevertheless, we believe the general feature-level and task-level source knowledge are also of great importance to unlabeled target domain.
Since the classifier may produce wrong predictions when tasks are difficult, the class-wise alignment cannot greatly favor learning discriminative features, and accordingly we need general knowledge transfer.
To maximally learn general knowledge, we want to align task-specific layer containing feature-level knowledge and classifier layer containing rich multimodal structure which is task-level knowledge.
However, we cannot align feature and class distribution separately, since the multimodal structures can only be captured sufficiently by the cross-covariance dependency between features and classes \cite{Hilbertspace}. To address this problem, \cite{JAN} proposes joint maximum mean discrepancy (JMMD) which can achieve cross-covariance dependency of feature representation and class prediction by multiplying interactions between them.
% Recent advances have revealed that, due to the domain shift across domains, even after the multi-layer feature abstraction in a deep network, the shifts still exist in the joint distributions $P_s(\x_s,y_s)$ and $P_t(\x_t,y_t)$ \cite{JAN}.
% We believe the features learned in task-specific layer are essential for domain knowledge transfer. Besides, the classifier output contains rich discriminative information that potentially reveals the multimode structures. By adapting them simultaneously, we can maximally learn general knowledge from both feature-level and task-level.
%
We therefore minimize the JMMD distance of multiple feature layers (task-specific layer and softmax layer) across domains to enable safe and effective knowledge transfer.
The key point is to control the importance of this part.
%DRCN manages to reduce the substantial cross-domain shift first by minimizing the JMMD metric of both domains, such that the general valuable source knowledge will be transferred to target successfully.

To be specific, we denote the activations of $n_s$ source data generated by the task-specific layer and softmax layer as $\{F_s({\x_s}_i)\}_{i=1}^{n_s}$ and $\{f_s({\x_s}_i)\}_{i=1}^{n_s}$, respectively. All the target data will go through the source task-specific layer and residual correction block simultaneously. The activations of $n_t$ target data generated by the element-wise summation and the softmax layer of source network as $\{F_t({\x_t}_j)\}_{j=1}^{n_t}$ and $\{f_s({\x_t}_j)\}_{j=1}^{n_t}$. To adapt the two layers effectively, we leverage the tensor product between feature-level and task-level activations to conduct joint distributions alignment.

%The feature fusion of ${\x_s}_i$ for the task-specific layer and softmax layer can be calculated as:
%\begin{align}
%\label{eq6:source-feature-fusion}
%{\ub_s}_i=F_s({\x_s}_i)\otimes f_s({\x_s}_i).
%\end{align}
%Similar definition can also be applied to target feature fusion of ${\x_t}_j$ as:
%\begin{align}
%\label{eq7:target-feature-fusion}
%{\ub_t}_j=F_t({\x_t}_j)\otimes f_s({\x_t}_j).
%\end{align}
Generally, the empirical estimate of JMMD between source and target domains in DRCN can be represented as:
\begin{footnotesize}
\begin{align}
\label{eq6:JMMD}
\mathcal{L}_{domain}&=\widehat{\mathcal{D}}_{\mathrm{JMMD}} (P_s,P_t)\nonumber\\
&=\frac{1}{n_s^2}\sum_{i,j=1}^{n_s}\kappa_1(F_s({\x_s}_i),F_s({\x_s}_j))\cdot \kappa_2(f_s({\x_s}_i),f_s({\x_s}_j))\nonumber\\
&-\frac{2}{n_sn_t}\sum_{i=1}^{n_s}\sum_{j=1}^{n_t}\kappa_1(F_s({\x_s}_i),F_t({\x_t}_j))\cdot \kappa_2(f_s({\x_s}_i),f_s({\x_t}_j))\nonumber\\
&+\frac{1}{n_t^2}\sum_{i,j=1}^{n_t}\kappa_1(F_t({\x_t}_i),F_t({\x_t}_j))\cdot \kappa_2(f_s({\x_t}_i),f_s({\x_t}_j)),
\end{align}
\end{footnotesize}
where $\kappa_1(\cdot,\cdot)$ and $\kappa_2(\cdot,\cdot)$ are the corresponding kernel functions. In DRCN, we choose Gaussian kernel function as the kernel function. In practical, to accelerate the computation speed of (\ref{eq6:JMMD}), we adopt a linear-time estimate of JMMD as follows:
\begin{footnotesize}
\begin{align}
\label{eq7:JMMD-linear}
&\mathcal{L}_{domain}= \nonumber\\ &\frac{2}{n}\sum_{i=1}^{n/2}\bigg(\kappa_1(F_s({\x_s}_{2i-1}),F_s({\x_s}_{2i}))\cdot \kappa_2(f_s({\x_s}_{2i-1}),f_s({\x_s}_{2i}))\nonumber\\
&~~~~~~~~~~~~~+\kappa_1(F_t({\x_t}_{2i-1}),F_t({\x_t}_{2i}))\cdot \kappa_2(f_s({\x_t}_{2i-1}),f_s({\x_t}_{2i}))\nonumber\\
&~~~~~~~~~~~~~-\kappa_1(F_s({\x_s}_{2i-1}),F_t({\x_t}_{2i}))\cdot \kappa_2(f_s({\x_s}_{2i-1}),f_s({\x_t}_{2i}))\nonumber\\
&~~~~~~~~~~~~~-\kappa_1(F_t({\x_t}_{2i-1}),F_s({\x_s}_{2i}))\cdot \kappa_2(f_s({\x_t}_{2i-1}),f_s({\x_s}_{2i}))\bigg).
\end{align}
\end{footnotesize}

To calculate (\ref{eq7:JMMD-linear}) during training, a minibatch of samples can be divided into quad-tuples, each containing two source samples and two target samples. This linear estimation of JMMD enables us to deal with large datasets effectively.

% Minimizing the difference of joint marginal and conditional distributions of source and target, only guarantees the source and target representations become more similar. Besides that, for partial domain adaptation, we also need to select out the irrelevant source subclasses, and assign larger importance scores to these most relevant source data. To achieve this goal, DRCN could automatically calculate the importance of each source class according to the target output probability distribution, and explicitly conduct weighted class-wise distribution matching to transfer the most related and useful knowledge from source to target.
%
\begin{figure}[tb]
\centering
\includegraphics[width=0.5\textwidth]{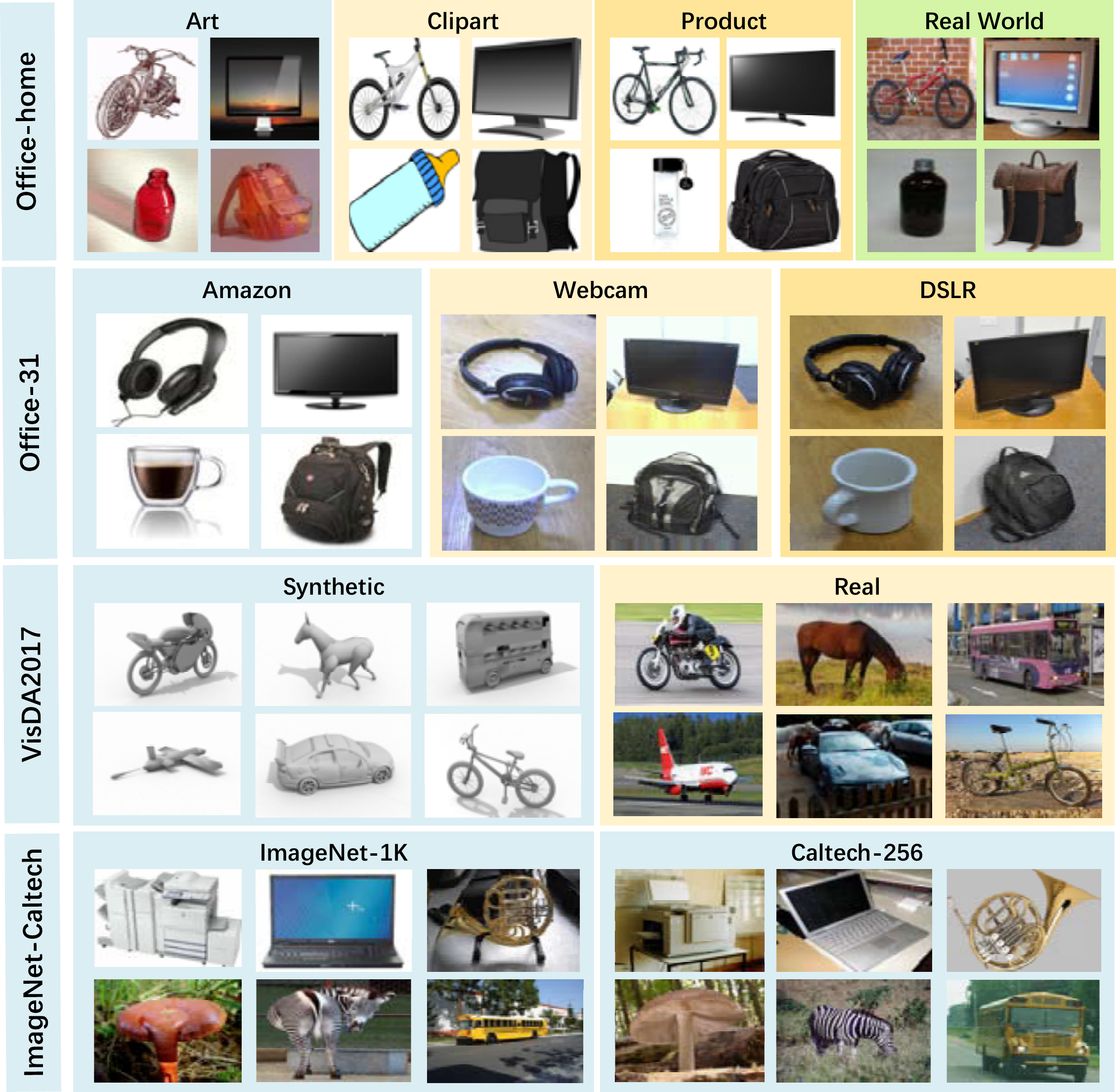}\vspace{-3mm}
\caption{Images examples of datasets Office-Home, Office-31, VisDA2017 and ImageNet-Caltech.}\vspace{-4mm}
\label{examples_imgs}
\end{figure}

\subsubsection{Overall Formulation of DRCN}

To enable effective partial domain adaptation, we propose a Deep Residual Correction Network (DRCN), which aligns the feature distributions across domains in the shared label space with larger importance scores, while jointly adapting general feature-level and task-level knowledge from source to target. By incorporating Eq. \eqref{eq4:source-loss}, \eqref{eq7:JMMD-linear} and \eqref{eq11:weighted-class-mmd}, the overall formulation of DRCN can be represented as:
\begin{align}
\label{eq12:DRCN-loss}
\mathcal{L}=\mathcal{L}_s+\alpha\mathcal{L}_{domain}+\beta\mathcal{L}_{class},
\end{align}
where $\alpha$ and $\beta$ are trade-offs to balance different terms.

\vspace{1mm}\noindent\textbf{Remark:} First, the soft labels of target data in DRCN could well characterize the output distribution of target data, which help identify the most relevant source subclasses effectively. We refer to DRCN with soft labels as \textbf{DRCN (soft label)}. Here, the soft labels in Eq. \eqref{eq8:class-weight} and Eq. \eqref{eq10:MKMMD} can also be replaced by hard labels, i.e., one-hot labels with the predicted class being 1. We define the hard label version as \textbf{DRCN (hard label)}. The pros and cons of the two schemes will be discussed in Section \ref{sec:experiment}.

Second, it is worthy to note that DRCN can not only deal with partial domain adaptation (target classes are a subset of source classes), but also address traditional domain adaptation scenarios (source and target label spaces are identical), by setting $\w$ to be an all one vector, which means each class weight is 1 with equal importance contribution for class-wise alignment loss. Moreover, DRCN can be naturally extended to solve fine-grained cross-domain visual recognition, where source and target are both fine-grained images, e.g., different models of car.

% by easily enlarging the value of $\alpha$ and wiping out the class-wise weights.
% More specifically, in traditional complete domain adaptation scenario, $\w$ is an all one vector, which means each class weight is 1 with equal importance contribution for class-wise alignment loss.
% Besides, The partial and complete domain adaptation results have demonstrated the universality of DRCN.

%\input{04-Experiment}
\section{Experiment}\label{sec:experiment}
In this section, we compare our proposed DRCN with several competitive unsupervised domain adaptation to verify the superiority of DRCN. Besides, we further conduct several empirical experiments to demonstrate the flexibility and effectiveness of our approach. Our source code is released as \url{https://github.com/wenqiwenqi1/DRCN}.

\begin{table}[tb]\footnotesize
  \centering
  \caption{Statistics of the benchmark datasets.}\vspace{-3mm}
    \begin{tabular}{|c|ccc|c|}
    \hline
    Dataset & Sub-domain & Abbr. & Sample & \#Class \bigstrut\\
    \hline
    \multirow{4}[2]{*}{Office-Home} & Art   & Ar    & 2427  & \multirow{4}[2]{*}{65} \bigstrut[t]\\
          & Clipart & Cl    & 4365  &  \\
          & Product & Pr    & 4439  &  \\
          & Real-World & Rw    & 4357  &  \bigstrut[b]\\
    \hline
    \multirow{3}[1]{*}{Office-31} & Amazon & A     & 2817  & \multirow{3}[1]{*}{31} \bigstrut[t]\\
          & DSLR  & D     & 498   &  \\
          & Webcam & W     & 795   &  \bigstrut[b]\\
    \hline
    \multirow{2}[2]{*}{VisDA2017} & Synthetic & S     & 152397 & \multirow{2}[2]{*}{12} \bigstrut[t]\\
          & Real  & R     & 55388 &  \bigstrut[b]\\
    \hline
    \multirow{2}[2]{*}{ImageNet-Caltech} & Caltech-256 & C     & 39037 & 256 \bigstrut[t]\\
          & ImageNet-1K & I    & 1285367 & 1000 \bigstrut[b]\\
    \hline
    \end{tabular}%
  \label{tab:statistics-dataset}\vspace{-5mm}
\end{table}%

We conduct the performance evaluation by using four cross-domain object recognition datasets: \textbf{Office-Home}\cite{Office-Home}, \textbf{Office-31}\cite{Office31}, \textbf{VisDA2017} and \textbf{ImageNet-Caltech}. Detailed information about these datasets is summarized in Table \ref{tab:statistics-dataset} and sample images per dataset are visualized in Figure \ref{examples_imgs}.

\textbf{Office-Home} is released at CVPR'17, containing 65 different objects from 4 domains, as: Artistic images (\textbf{Ar}), Clip Art (\textbf{Cl}), Product images (\textbf{Pr}) and Real-World images (\textbf{Rw}). For each domain, the dataset consists of images found typically in Office and Home settings. Specifically, the images of domain \textbf{Ar} are paintings, sketches or artistic depictions, and that of \textbf{Cl} are clipart images. \textbf{Rw} comprises regular images captured by cameras, while the images in \textbf{Pr} are all without background. These images in the dataset are crawled through several search engines and online image directories, and widely used in domain adaptation algorithms \cite{PADA,ARCPIP}. For these four domains, similar to \cite{PADA}, the first 25 classes (in alphabetical order) in that domain are taken as target categories, if one of them is selected as target domain. Considering all the domain combinations, we build 12 cross-domain learning tasks: \textbf{Ar}$\rightarrow$\textbf{Cl}, ..., \textbf{Rw}$\rightarrow$\textbf{Pr}.

%: \textbf{Ar}$\rightarrow$\textbf{Cl}, \textbf{Ar}$\rightarrow$\textbf{Pr}, \textbf{Ar}$\rightarrow$\textbf{Rw}, \textbf{Cl}$\rightarrow$\textbf{Ar}, \textbf{Cl}$\rightarrow$\textbf{Pr}, \textbf{Cl}$\rightarrow$\textbf{Rw}, \textbf{Pr}$\rightarrow$\textbf{Ar}, \textbf{Pr}$\rightarrow$\textbf{Cl}, \textbf{Pr}$\rightarrow$\textbf{Rw}, \textbf{Rw}$\rightarrow$\textbf{Ar}, \textbf{Rw}$\rightarrow$\textbf{Cl}, \textbf{Rw}$\rightarrow$\textbf{Pr}.
%

\textbf{Office-31} is widely adopted by deep transfer learning methods \cite{DAN-PAMI,ARCPIP,SAN}, with 4110 images that are commonly encountered objects in office settings, such as monitors, bottles and backpacks. It includes 31 classes and involves three distinct domains, as Amazon (\textbf{A}, which is downloaded from online merchants), Webcam (\textbf{W}) and DSLR (\textbf{D}). \textbf{W}, \textbf{D} contain images taken by web cameras and digital SLR cameras, respectively. We follow similar settings as \cite{SAN,PADA} and select 10 classes shared by Office-31 and Caltech-256 \cite{caltech} as target categories. The corresponding images in Office-31 are defined as target domain. Hence, we can construct 6 domain adaptation transfer tasks: \textbf{A}$\rightarrow$\textbf{W}, ..., \textbf{D}$\rightarrow$\textbf{W}.
%: \textbf{A}$\rightarrow$\textbf{W}, \textbf{A}$\rightarrow$\textbf{D}, \textbf{W}$\rightarrow$\textbf{A}, \textbf{W}$\rightarrow$\textbf{D}, \textbf{D}$\rightarrow$\textbf{A}, \textbf{D}$\rightarrow$\textbf{W}.
%

\textbf{VisDA2017} is a large-scale dataset for cross-domain object classification which was first present in 2017 Visual Domain Adaptation (VisDA) Challenge. In the experiment, we use the training and validation images provided by the competition as 2 domains: one comprises synthetic 2D renderings of 3D models generated from different angles, and the other consists photo-realistic images or real images. Both of them have 12 categories in common. We denote the domain with synthetic images as \textbf{S}, while the domain with real images as \textbf{R}. Besides, we assume that if one domain is selected as target domain, the first 6 categories (in alphabetical order) will be chosen as target categories with corresponding images as target domain. Hence, we can conclude 2 transfer tasks: \textbf{S}$\rightarrow$\textbf{R} and \textbf{R}$\rightarrow$\textbf{S}.
% Table generated by Excel2LaTeX from sheet 'partial-home'
% greatly improve classification accuracy for task
\begin{table*}[tb]\footnotesize
  \centering
  \caption{Accuracy (\%) on Office-Home for \textbf{partial} transfer learning tasks (ResNet-50).}\vspace{-3mm}
    \setlength{\tabcolsep}{1.0mm}{
    \begin{tabular}{|c|ccccccccccccc|}
    \hline
    \multirow{2}[4]{*}{Method} & \multicolumn{13}{c|}{Office-Home} \bigstrut\\
\cline{2-14}          & Ar$\rightarrow$Cl  & Ar$\rightarrow$Pr  & Ar$\rightarrow$Rw  & Cl$\rightarrow$Ar  & Cl$\rightarrow$Pr  & Cl$\rightarrow$Rw  & Pr$\rightarrow$Ar  & Pr$\rightarrow$Cl  & Pr$\rightarrow$Rw  & Rw$\rightarrow$Ar  & Rw$\rightarrow$Cl  & Rw$\rightarrow$Pr  & Average \bigstrut\\
    \hline
    ResNet\cite{resnet} & 38.6  & 60.8  & 75.2  & 39.9  & 48.1  & 52.9  & 49.7  & 30.9  & 70.8  & 65.4  & 41.8  & 70.4  & 53.7  \bigstrut[t]\\
    DAN\cite{DAN}   & 44.4  & 61.8  & 74.5  & 41.8  & 45.2  & 54.1  & 46.9  & 38.1  & 68.4  & 64.4  & 45.4  & 68.9  & 54.5  \\
    RevGrad\cite{DA_bp}  & 44.9  & 54.1  & 69.0  & 36.3  & 34.3  & 45.2  & 44.1  & 38.0  & 68.7  & 53.0  & 34.7  & 46.5  & 47.4  \\
    RTN\cite{RTN}   & 49.4  & 64.3  & 76.2  & 47.6  & 51.7  & 57.7  & 50.4  & 41.5  & 75.5  & 70.2  & 51.8  & 74.8  & 59.3  \\
    IWAN\cite{WAN-for-partial}  & 53.9  & 54.5  & 78.1  & 61.3  & 48.0 & 63.3  & 54.2  & \textbf{52.0} & \textbf{81.3} & 76.5  & 56.8  & \textbf{82.9} & 63.6  \\
    SAN\cite{SAN}  & 44.4 & 68.7  & 74.6  & \textbf{67.5} & 65.0  & \textbf{77.8} & 59.8  & 44.7 & 80.1  & 72.2  & 50.2  & 78.7  & 65.3  \\
   % PADA-classifier & 47.5  & 58.2  & 74.3  & 43.6  & 37.9  & 51.9  & 48.2  & 41.7  & 71.6  & 67.1  & 53.0  & 71.6  & 55.6  \\
%    PADA-adversarial & 47.1  & 47.5  & 67.5  & 41.3  & 39.7  & 52.7  & 43.1  & 35.9  & 70.5  & 61.8  & 48.2  & 70.1  & 52.1  \\
    PADA\cite{PADA}  & 52.0  & 67.0  & 78.7  & 52.2  & 53.8  & 59.0  & 52.6  & 43.2  & 78.8  & 73.7  & 56.6  & 77.1  & 62.1  \bigstrut[b]\\
   \hline
    \textbf{DRCN (soft label)} & \textbf{54.0} & \textbf{76.4} & \textbf{83.0}  & 62.1  & 64.5  & 71.0  & \textbf{70.8} & 49.8  & 80.5  & \textbf{77.5}  & \textbf{59.1}  & 79.9  & \textbf{69.0} \bigstrut[t]\\
    \textbf{DRCN (hard label)} & 51.6  & 75.8  & 82.0 & 62.9  & \textbf{65.1} & 72.9  & 67.4  & 50.0  & 81.0  & 76.4 & 57.7  & 79.3  & 68.5 \bigstrut[b]\\
    \hline
    \end{tabular}}
  \label{tab:partial-home}\vspace{-3mm}
\end{table*}%

\textbf{ImageNet-Caltech} is a challenging dataset which consists of ImageNet-1K and Caltech-256 and we denote them as \textbf{I} and \textbf{C} respectively. Comparing with former datasets, it is larger with over 1M images across 1000 classes in ImageNet-1K (\textbf{I}) and over 39K images across 256 classes in Caltech-256 (\textbf{C}).
% classes in which ImageNet has 1000 classes and Caltech has 256 classed with sharing 84 ones.
% it contains over 1M images across 1000 classes in ImageNet and 39K images across 256 classes in Caltech, with sharing 84 classes.
Since there are 84 common classes in both, we therefore build two partial transfer tasks: \textbf{I (1000)} $\rightarrow$ \textbf{C (84)} and \textbf{C (256)} $\rightarrow$ \textbf{I (84)}. And similar to \cite{PADA}, we use ImageNet-1K validation set as target domain when it comes to task \textbf{C (256)} $\rightarrow$ \textbf{I (84)}.

% \subsubsection{ImageCLEF-DA}
% It is a benchmark for ImageCLEF 2014 challenge, containing 4 different domains which are selected from other 4 datasets: Caltech-256 (\textbf{C}), ImageNet ILSVRC 2012 (\textbf{I}), Pascal VOC 2012 (\textbf{P}) and Bing (\textbf{B}). These domains are shared with 12 common categories and each category has 50 images, therefore there are 600 images in each domain.

% Specifically, we denote the four domains above as C, I, P, B and when a doamin is used as target domain, we will choose (in alphabetical order) its first 6 classes as target categories and all images under these classes as target domain. Considering all domain combinations, we can build 12 transfer tasks, i.e., \textbf{C}$\rightarrow$\textbf{I}, \textbf{C}$\rightarrow$\textbf{P}, \textbf{C}$\rightarrow$\textbf{B}, \textbf{I}$\rightarrow$\textbf{C}, \textbf{I}$\rightarrow$\textbf{P}, \textbf{I}$\rightarrow$\textbf{B}, \textbf{P}$\rightarrow$\textbf{C}, \textbf{P}$\rightarrow$\textbf{I}, \textbf{P}$\rightarrow$\textbf{B}, \textbf{B}$\rightarrow$\textbf{C}, \textbf{B}$\rightarrow$\textbf{I}, \textbf{B}$\rightarrow$\textbf{P}.
%
\subsection{Baseline Methods}
For the diversity of the experiment, we first take two shallow methods: GFK\cite{GFK}, TCA\cite{TCA} with the deep features of ResNet-50 layer \textit{pool5} as baselines. To extensively verify the effectiveness of DRCN, we compare DRCN with 12 traditional or partial deep domain adaptation methods: ResNet\cite{resnet}, DAN \cite{DAN}, RevGrad \cite{DA_bp}, RTN \cite{RTN}, JAN \cite{JAN}, LEL \cite{LEL}, ADDA \cite{ADDA}, MADA\cite{MADA}, CDAN \cite{CDAN}, SAN\cite{SAN}, IWAN\cite{WAN-for-partial}, and PADA\cite{PADA}. Note that several results are directly from the published papers if we follow the same setting.

\subsection{Setup}
We follow standard protocols as \cite{DAN}, in which the source data are all labeled while target data are unlabeled. Besides, we average the accuracy of each cross-domain task by performing three random experiments and resize the input images to 256 $\times$ 256.

For all the experiments, DRCN and other compared deep domain adaptation methods are implemented with PyTorch, and fine-tuned from PyTorch-provided model of ResNet-50 pre-trained on ImageNet similar to the previous works \cite{JAN,PADA}. In this paper, we use the same learning rate as JAN\cite{JAN} in classifier layer, which is 10 times larger than other layers. Since the residual layer is trained from scratch and it needs to be very precise, we set its learning rate to be one tenth that of the other layers. In addition, we use stochastic gradient descent (SGD) with momentum of 0.9 and a learning rate annealing strategy as \cite{PADA}. We compute the class weights after each epoch when all target samples are trained. Therefore, the weights would be updated dynamically. For the fair comparison, we use the importance weighted cross-validation technique as \cite{CDAN} to select hyper-parameters, and set $\alpha = 0.1$ and $\beta = 0.05$ throughout all the experiments. Moreover, we will give parameter sensitivity analysis for DRCN, which indicates that for reasonable parameter values, DRCN always can achieve stable performance.
% Considered with the trade-off between JMMD and conditional probability in our method (DRCN)
%
\subsection{Results and Analysis}
We evaluate DRCN on the following four datasets, Office-Home, Office-31, VisDA2017 and ImageNet-Caltech, by comparing with competitive traditional and partial deep domain adaptation methods, and make some discussions.
\subsubsection{Partial Transfer Results on Office-Home}
%Table \ref{tab:partial-home} summarizes the experimental results of Office-Home, in which we can see that DRCN outperforms all comparison methods on all tasks. It is noteworthy that on some easy domain transfer tasks, where the source domain and target domain are similar, i.e., \textbf{Ar$\rightarrow$Rw}, \textbf{Pr$\rightarrow$Rw}, \textbf{Rw$\rightarrow$Ar}, \textbf{Rw$\rightarrow$Pr}, DRCN achieves comparable performance on these tasks, and promotes significantly on other hard domain transfer tasks, i.e., \textbf{Pr$\rightarrow$Ar}, and \textbf{Cl$\rightarrow$Rw}, in which domain shifts are remarkably different. Besides, hard label based DRCN is relatively better than soft label based DRCN in terms of the accuracy. Note that some traditional domain adaptation methods, like DAN, RevGrad, have worse performances on most tasks compared with ResNet. And PADA beats the baseline ResNet by an average accuracy improvement of 8.4\%. Excitingly, our approach is \textbf{4.5\%} and \textbf{2.8\%} better than PADA in terms of soft label and hard label, respectively.
Table \ref{tab:partial-home} summarizes the experimental results of Office-Home, in which we can observe that DRCN outperforms all comparison methods on most tasks. Note that some traditional domain adaptation methods, like DAN, RevGrad, achieve worse performance on most tasks compared with ResNet, which implies that only conducting distribution alignment across domains is not enough for partial domain adaptation problems. Whereas PADA beats ResNet by an average accuracy improvement of 8.4\% while IWAN and SAN increase 1.5\% and 3.2\% respectively compared to PADA. It is reasonable since IWAN and SAN consider to alleviate negative transfer by detecting outlier classes with various weighting schemes. It is worthy to note that DRCN gains the highest accuracies on 7 out of 12 tasks, and promotes significantly on several domain transfer tasks, i.e., \textbf{Ar$\rightarrow$Pr}, \textbf{Pr$\rightarrow$Ar} and \textbf{Ar$\rightarrow$Rw}. For average accuracy, DRCN works best by outperforming baseline SAN with at least \textbf{3.2\%} improvement, indicating that DRCN is more conductive to weaken the influence from the irrelevant source classes thereby encouraging relevant information transfer among shared classes.
%Although SAN and IWAN show best performance on some tasks, in which DRCN show sightly lower accuracy, our approach could achieve larger improvements on hard tasks.
%Besides, hard label based DRCN is relatively better than soft label based DRCN in terms of the accuracy.
Besides, soft label based DRCN is slightly better than hard label based DRCN, because soft label contains class structure information that can help correct prediction deviations.
%since hard label based approach is able to avoid assigning negative weights as much as possible on those irrelevant classes, since its one-hot encoding attribute.
%our approach can significantly discriminate outlier classes from source domain in Office-Home dataset.}
%Excitingly, our approach is \textbf{6.7\%} and \textbf{6.3.8\%} better than PADA in terms of soft label and hard label, respectively.}
% Table generated by Excel2LaTeX from sheet 'partial-31'
\begin{table}[tb]\scriptsize
 % \centering
% \vspace{0.5em}
  \caption{Accuracy (\%) on Office-31 for \textbf{partial} transfer learning tasks (ResNet-50).}\vspace{-3mm}
    \setlength{\tabcolsep}{1mm}{
    \begin{tabular}{|c|ccccccc|}
    \hline
    \multirow{2}[4]{*}{Method} & \multicolumn{7}{c|}{Office-31} \bigstrut\\
\cline{2-8}          & A$\rightarrow$W    & D$\rightarrow$W    & W$\rightarrow$D    & A$\rightarrow$D    & D$\rightarrow$A    & W$\rightarrow$A    & Average \bigstrut\\
    \hline
    ResNet\cite{resnet} & 54.5  & 94.6  & 94.3  & 65.6  & 73.2  & 71.7  & 75.6  \bigstrut[t]\\
    DAN\cite{DAN}   & 46.4  & 53.6  & 58.6  & 42.7  & 65.7  & 65.3  & 55.4  \\
    RevGrad\cite{DA_bp} & 41.4  & 46.8  & 38.9  & 41.4  & 41.3  & 44.7  & 42.4  \\
    ADDA\cite{ADDA}  & 43.7  & 46.5  & 40.1  & 43.7  & 42.8  & 46.0  & 43.8  \\
    RTN\cite{RTN}   & 75.3  & 97.1  & 98.3  & 66.9  & 85.6  & 85.7  & 84.8  \\
    JAN\cite{JAN}   & 43.4  & 53.6  & 41.4  & 35.7  & 51.0  & 51.6  & 46.1  \\
    LEL\cite{LEL}   & 73.2  & 93.9  & 96.8  & 76.4  & 83.6  & 84.8  & 84.8  \\
    IWAN\cite{WAN-for-partial}  & 89.2  & 99.3  & 99.4  & 90.5  & \textbf{95.6}  & 84.7  & 93.1 \\
    SAN\cite{SAN}   & \textbf{93.9}  & 99.3  & 99.4  & \textbf{94.3}  & 94.2  & 88.7  & 95.0 \\
  %  PADA-classifier & 83.1  & 99.3  & 100.0  & 80.2  & 90.1  & 92.3  & 90.9  \\
%    PADA-adversarial & 65.8  & 97.3  & 97.5  & 77.1  & 87.3  & 87.4  & 85.4  \\
    PADA\cite{PADA}  & 86.5  & 99.3  & \textbf{100.0}  & 82.2  & 92.7  & 95.4  & 92.7  \bigstrut[b]\\
    \hline
    \textbf{DRCN (soft label)} & 88.5  & \textbf{100.0}  & \textbf{100.0}  & 86.0  & \textbf{95.6}  & \textbf{95.8}  & 94.3  \bigstrut[t]\\
    \textbf{DRCN (hard label)} &90.8  & \textbf{100.0}  & \textbf{100.0}  & \textbf{94.3} & 95.2  & 94.8  & \textbf{95.9}  \bigstrut[b]\\
    \hline
    \end{tabular}}
  %  }%
  \label{tab:partial-31}\vspace{-5mm}
\end{table}%
\subsubsection{Partial Transfer Tasks Results on Office-31}

The classification results of Office-31 are reported in Table \ref{tab:partial-31}, from which we obviously observe that DRCN continuously outperforms other methods, however, by a small margin, e.g., average accuracy \textbf{94.3\%} in soft label while \textbf{95.9\%} in hard label, and SAN achieves the second highest accuracy of 95.0\% followed by IWAN with 93.1\%, while PADA gains 92.7\%.
It is clear that hard label based DRCN achieves comparable results while soft label based DRCN performs slightly worse, especially in task A$\rightarrow$D where hard label based DRCN gets \textbf{8.3\%} improvement compared to soft label based DRCN.
This is because soft label based approach is likely to assign the weights to some unrelated classes, resulting in poor results, when classifier can accurately make prediction in a large probability.
%because Office-31 does not have such a large discriminative difference between domains as Office-Home, resulting in equal performance among advanced methods, and weak results of soft label based DRCN due to similar classes' interference.
Additionally, we can get some interesting observations that DAN, RevGrad, ADDA and JAN all perform much worse than ResNet on most of tasks, by 6.4\% to 55.4\% descent, demonstrating that partial transfer task is a complex scenario which can not be easily tackled by traditional domain adaptation methods.
\subsubsection{Partial Transfer Results on VisDA2017 and ImageNet-Caltech}

In contrast to previous datasets, VisDA2017 and ImageNet-Caltech have much larger domain scales. However, DRCN can substantially improve the accuracy and is comparable to PADA, IWAN and SAN based on partial domain adaptation as shown in Table \ref{tab:partial-visda}.
%We can clearly find out that ResNet, DAN, RTN, RevGrad and PADA obtain better performance.
Although PADA achieves the highest accuracy of 76.5\% which is better than our best result of 74.2\% on task \textbf{R$\rightarrow$S}, but PADA is worse than DRCN on others. Especially for task \textbf{S$\rightarrow$R}, DRCN significantly outperforms PADA by an increase of at least \textbf{3.7\%}.

On the perspective of ImageNet-Caltech, our method promotes the accuracy in task \textbf{C$\rightarrow$I} by a large margin, and gains comparable result in task \textbf{I$\rightarrow$C}. We report the average accuracy on VisDA2017 and ImageNet-Caltech, manifesting that our method outperforms other baselines. Besides, soft label based DRCN has higher average accuracy than that of hard label based method, which shows that in a large-scale dataset, soft label has a greater impact in enhancing the contributions of relevant classes, and down-weighing contributions of irrelevant ones.
\vspace{-3mm}
%In contrast to previous datasets, VisDA2017 has a much larger domain size, however, DRCN can still improve the accuracy substantially and comparable to PADA based on partial domain adaptation as shown in Table \ref{tab:partial-visda}. We can clearly find out that ResNet, DAN, RTN, RevGrad and PADA are getting better on the tasks. Although PADA achieves highest accuracy of 76.5\%  which is better than our best result of 74.2\%  in task \textbf{R$\rightarrow$S}, but PADA is slightly worse than our method on the whole, especially on task \textbf{S$\rightarrow$R}, where our method significantly outperforms PADA by an increase of at least \textbf{3.7\%}.
% Table generated by Excel2LaTeX from sheet 'complete-office-31'
% Table generated by Excel2LaTeX from sheet 'partial-visda'
\begin{table}[htbp]\footnotesize
 \centering
  \caption{Accuracy (\%) on VisDA2017 and ImageNet-Caltech for \textbf{partial} transfer learning tasks (ResNet-50).}\vspace{-3mm}
    %\begin{tabular}{|c|@{}>{\hfil}p{11.5mm}<{\hfil}@{}>{\hfil}p{9mm}<{\hfil}|@{}>{\hfil}p{13mm}<{\hfil}@{}>{\hfil}p{11.5mm}<{\hfil}|p{7mm}<{\centering}|}
    \setlength{\tabcolsep}{1.5mm}{
    \begin{tabular}{|c|cc|@{}>{\hfil}p{13mm}<{\hfil}@{}>{\hfil}p{11.5mm}<{\hfil}|c|}
    \hline
    \multirow{2}[3]{*}{Method} & \multicolumn{2}{c|}{VisDA2017} & \multicolumn{2}{c|}{ImageNet-Caltech} & \multirow{2}[3]{*}{Average} \bigstrut\\
\cline{2-5}          & R$\rightarrow$S    & S$\rightarrow$R    & I$\rightarrow$C    & C$\rightarrow$I    &  \bigstrut\\
    \hline
    ResNet\cite{resnet} & 64.3  & 45.3  & 69.7  & 71.3  & 62.6  \bigstrut[t]\\
    DAN\cite{DAN}   & 68.4  & 47.6  & 71.3  & 60.1  & 61.9  \\
    RevGrad\cite{DA_bp} & 73.8  & 51.0  & 70.8  & 67.7  & 65.8  \\
    RTN\cite{RTN}   & 72.9  & 50.0  & 75.5  & 66.2  & 66.2  \\
    IWAN\cite{WAN-for-partial}  & 71.3   & 48.6   & \textbf{78.1}  & 73.3  & 67.8 \\
    SAN\cite{SAN}   & 69.7  & 49.9  &77.8  & 75.3  & 68.2  \\
    PADA\cite{PADA}  & \textbf{76.5}  & 53.5  & 75.0  & 70.5  & 68.9  \bigstrut[b]\\
    \hline
    \textbf{DRCN (soft label)} & 73.2  & \textbf{58.2}  &75.3  & \textbf{78.9}  & \textbf{71.4}  \bigstrut[t]\\
    \textbf{DRCN (hard label)} & 74.2  & 57.2  & 74.1  & 77.5  & 70.8  \bigstrut[b]\\
    \hline
    \end{tabular}}
  \label{tab:partial-visda}
\end{table}%

\vspace{-3mm}
\textbf{Summary}: First, from the results of Tables \ref{tab:partial-home} and \ref{tab:partial-31}, we notice that traditional transfer learning methods cannot consistently outperform standard ResNet, which is extremely obvious in Table \ref{tab:partial-31}. However, some approaches on partial domain adaptation tasks can easily improve the accuracy by huge margins. This validates that partial transfer adaptation is a much more challenging problem for traditional domain adaptation methods. These methods only consider the general cross-domain features, which would lead to negative transfer.

Second, soft label based approach tends to work well in large-scale datasets (VisDA2017 and ImageNet-Caltech) while hard label based approach works well in small-scale datasets (Office-31 and Office-Home). This is because that classifier is prone to predict wrong label in large-scale dataset. In this case, hard label will falsely transfer unrequired class information, whereas soft label brings additional information from other classes which may cover desirable knowledge and then mitigates prediction deviations.
However, if classifier works on small-scale datasets and produces correct prediction, it is detrimental to assign weights on those irrelevant classes. This problem can be avoided by hard label as its one-hot encoding attribute.
%However,  is likely to assign the weights to some unrelated classes, resulting in poor results, when we cannot make sure which subclasses belong to the target domain and the classifier cannot accurately identify the image. In contrast, hard label based approach is able to avoid assigning negative weights as much as possible on those irrelevant classes, since its one-hot encoding attribute.

Third, from small-scale to large-scale datasets, DRCN achieves significant improvements on most tasks. Those encouraging results indicate that DRCN can effectively identify the most relevant source classes and learn more transferable features for partial domain adaptation.

\subsection{Extension: Traditional Transfer Learning Tasks}

As stated in Section \ref{sec:method}, by enlarging the value of $\alpha$ and setting the class-wise weights equal to 1, i.e. $\w=\mathbf{1}$, DRCN can be easily generalized to deal with traditional domain adaptation scenarios. To evaluate the generalization ability, we extensively compare DRCN with several popular traditional domain adaptation methods including DAN \cite{DAN}, RevGrad \cite{DA_bp}, RTN \cite{RTN}, JAN \cite{JAN}, ADDA \cite{ADDA}, MADA\cite{MADA}, CDAN \cite{CDAN}, with two datasets on traditional transfer tasks, where the source domain and target domain have identical label spaces. For traditional transfer tasks, we set $\alpha=1.5$ and $\beta=0.05$ as default. The results are shown in Tables \ref{tab:traditional-home} and \ref{tab:traditional-31}. Additionally, in the traditional domain adaptation scenario, DRCN could also be used to address the problems of the fine-grained recognition in the wild.
\begin{table*}[tb]\footnotesize
  \centering
  \caption{Accuracy (\%) on Office-Home for \textbf{traditional} transfer learning tasks (ResNet-50).}\vspace{-3mm}
    \setlength{\tabcolsep}{1.0mm}{
    \begin{tabular}{|c|ccccccccccccc|}
    \hline
    \multirow{2}[4]{*}{Method} & \multicolumn{13}{c|}{Office-Home} \bigstrut\\
\cline{2-14}   &Ar$\rightarrow$Cl&Ar$\rightarrow$Pr&Ar$\rightarrow$Rw& Cl$\rightarrow$Ar  & Cl$\rightarrow$Pr  & Cl$\rightarrow$Rw  & Pr$\rightarrow$Ar  & Pr$\rightarrow$Cl  & Pr$\rightarrow$Rw  & Rw$\rightarrow$Ar  & Rw$\rightarrow$Cl  & Rw$\rightarrow$Pr  & Average \bigstrut\\
    \hline
    ResNet\cite{resnet} &34.9& 50.0& 58.0& 37.4  & 41.9  & 46.2  & 38.5  & 31.2  & 60.4  & 53.9  & 41.2  & 59.9  & 46.1  \bigstrut[t]\\
    DAN\cite{DAN}   &43.6& 57.0& 67.9& 45.8  & 56.5  & 60.4  & 44.0  & 43.6  & 67.7  & 63.1  & 51.5  & 74.3  & 56.3  \\
    RevGrad\cite{DA_bp} &45.6& 59.3&70.1& 47.0  & 58.5  & 60.9  & 46.1  & 43.7  & 68.5  & 63.2  & 51.8  & 76.8  & 57.6  \\
    JAN\cite{JAN}   &45.9& 61.2& 68.9& 50.4  & 59.7  & 61.0  & 45.8  & 43.4  & 70.3  & 63.9  & 52.4  & 76.8  & 58.3  \\
    CDAN\cite{CDAN}  & 49.0  & 69.3  &74.5  & 54.4  & 66.0  & 68.4  & 55.6  & 48.3  & 75.9  & 68.4  & 55.4 & 80.5  & 63.8  \\
    CDAN+E\cite{CDAN} & \textbf{50.7} & 70.6  & 76.0  & 57.6  & \textbf{70.0}  & 70.0  &57.4  & \textbf{50.9} & \textbf{77.3} & 70.9  & \textbf{56.7} & \textbf{81.6} & 65.8  \bigstrut[b]\\
    \hline
    \textbf{DRCN (soft label)} & 50.6  & \textbf{72.4} & \textbf{76.8} & \textbf{61.9} & 69.5  & \textbf{71.3} & \textbf{60.4} & 48.6  & 76.8  & 72.9  & 56.1  & 81.4  & \textbf{66.6}  \bigstrut[t]\\
    \textbf{DRCN (hard label)} & 49.1  & 71.2  & 76.1  & 61.7  & \textbf{70.0} & 71.1  & 59.4  & 49.1  & 76.8  & \textbf{73.3} & 55.9  & 80.8  & 66.2 \bigstrut[b]\\
    \hline
    \end{tabular}}
  \label{tab:traditional-home}\vspace{-3mm}
\end{table*}%
%
% Table generated by Excel2LaTeX from sheet 'complete-home'
\subsubsection{Traditional Transfer Results on Office-Home}

The results on the 12 transfer tasks of Office-Home are shown in Table \ref{tab:traditional-home}. We observe that DRCN outperforms all other methods on most tasks. Regarding those baseline methods, their average accuracy can be over 55\% which is much better than the baseline ResNet. Additionally, CDAN+E wins the highest accuracies among baseline methods, with 7.5\% and 2\% improvements compared to JAN and CDAN respectively.
For the average accuracy, we can observe that soft label based DRCN is better with hard label based DRCN on most tasks, and both of them can be competitive with CDAN+E. For example, on the task \textbf{Cl$\rightarrow$Ar}, our method has increased by \textbf{4.3\%} and \textbf{4.1\%} respectively under soft label approach and hard label approach compared to CDAN+E. This suggests that our method can be successfully extended to address traditional transfer tasks.
\subsubsection{Traditional Transfer Tasks Results on Office-31}
Table \ref{tab:traditional-31} shows the classification accuracy results on Office-31. It is clearly that TCA and GFK are comparable to each other, while RevGrad and ADDA take slightly advantages over DAN and RTN. Similarly, MADA beats JAN by an increase of 0.9\%. The average classification accuracy of DRCN of soft label on all 6 transfer tasks is \textbf{87.2\%} and that of hard label on these tasks is \textbf{87.0\%}, in which we can see that the best accuracy increase is 0.7\% against CDAN. Although our approach is slightly worse than the best baseline CDAN+E by a marginal decrease of 0.5\%, DRCN still can achieve accuracy improvements on some hard tasks, such as \textbf{D$\rightarrow$A} and \textbf{W$\rightarrow$A}, and attain comparable results on other easy tasks, indicating that DRCN also has potential power to substantially improve the classification accuracies in traditional domain adaptation scenarios.

\textbf{Summary}:
First, as it can be seen in Table \ref{tab:traditional-home} and \ref{tab:traditional-31}, DRCN (soft label) is slightly better than DRCN (hard label), which means the target prediction distribution could provide more discriminative knowledge comparing to one-hot pseudo target labels. Thus, it results in better performance when encountering traditional domain adaptation scenarios.

Second, by adjusting the values of $\alpha$ and setting $\w=\mathbf{1}$, we can get comparable or even better results against other competitive domain adaptation methods. Those convincing results on challenging tasks show that the modified DRCN can also adapt to traditional domain adaptation problem.

Third, comparing with RTN \cite{RTN}, DRCN could achieve much higher prediction accuracies, which indicates mitigating the feature discrepancy between domains rather than classifier difference is more crucial for addressing domain adaptation problems.

\begin{table}[htbp]\scriptsize
  \centering
  \caption{Accuracy (\%) on Office-31 for \textbf{traditioanl} transfer learning tasks (ResNet-50).}\vspace{-3mm}
      \setlength{\tabcolsep}{1mm}{
    \begin{tabular}{|c|ccccccc|}
    \hline
    \multirow{2}[4]{*}{Method} & \multicolumn{7}{c|}{Office-31} \bigstrut\\
\cline{2-8}          & A$\rightarrow$W    & D$\rightarrow$W    & W$\rightarrow$D    & A$\rightarrow$D    & D$\rightarrow$A    & W$\rightarrow$A    & Average \bigstrut\\
    \hline
    ResNet\cite{resnet} & 68.4  & 96.7  & 99.3  & 68.9  & 62.5  & 60.7  & 76.1  \bigstrut[t]\\
    TCA\cite{TCA}   & 72.7  & 96.7  & 99.6  & 74.1  & 61.7  & 60.9  & 77.6  \\
    GFK\cite{GFK}   & 72.8  & 95.0  & 98.2  & 74.5  & 63.4  & 61.0  & 77.5  \\
    DAN\cite{DAN}   & 80.5  & 97.1  & 99.6  & 78.6  & 63.6  & 62.8  & 80.4  \\
    RTN\cite{RTN}   & 84.5  & 96.8  & 99.4  & 77.5  & 66.2  & 64.8  & 81.6  \\
    RevGrad\cite{DA_bp} & 82.0  & 96.9  & 99.1  & 79.7  & 68.2  & 67.4  & 82.2  \\
    ADDA\cite{ADDA}  & 86.2  & 96.2  & 98.4  & 77.8  & 69.5  & 68.9  & 82.9  \\
    JAN\cite{JAN}   & 85.4  & 97.4  & 99.8  & 84.7  & 68.6  & 70.0  & 84.3  \\
    MADA\cite{MADA}  & 90.0  & 97.4  & 99.6  & 87.8  & 70.3  & 66.4  & 85.2  \\
    CDAN\cite{CDAN}  & 93.1 & 98.2  & \textbf{100.0}  & 89.8  & 70.1  & 68.0  & 86.5  \\
    CDAN+E\cite{CDAN} & \textbf{94.1} & \textbf{98.6} & \textbf{100.0} & \textbf{92.9} & 71.0 & 69.3  & \textbf{87.7} \bigstrut[b]\\
    \hline
    \textbf{DRCN (soft label)} & 93.1  & 98.0  & \textbf{100.0} & 89.4  & 71.4  & \textbf{71.0}  & 87.2 \bigstrut[t]\\
    \textbf{DRCN (hard label)} & 92.7  & 97.6  & 99.8  & 88.4  & \textbf{72.7} & \textbf{71.0} & 87.0 \bigstrut[b]\\
    \hline
    \end{tabular}}
  \label{tab:traditional-31}
\end{table}%

\begin{figure}[!t]
\centering
\includegraphics[width=0.48\textwidth]{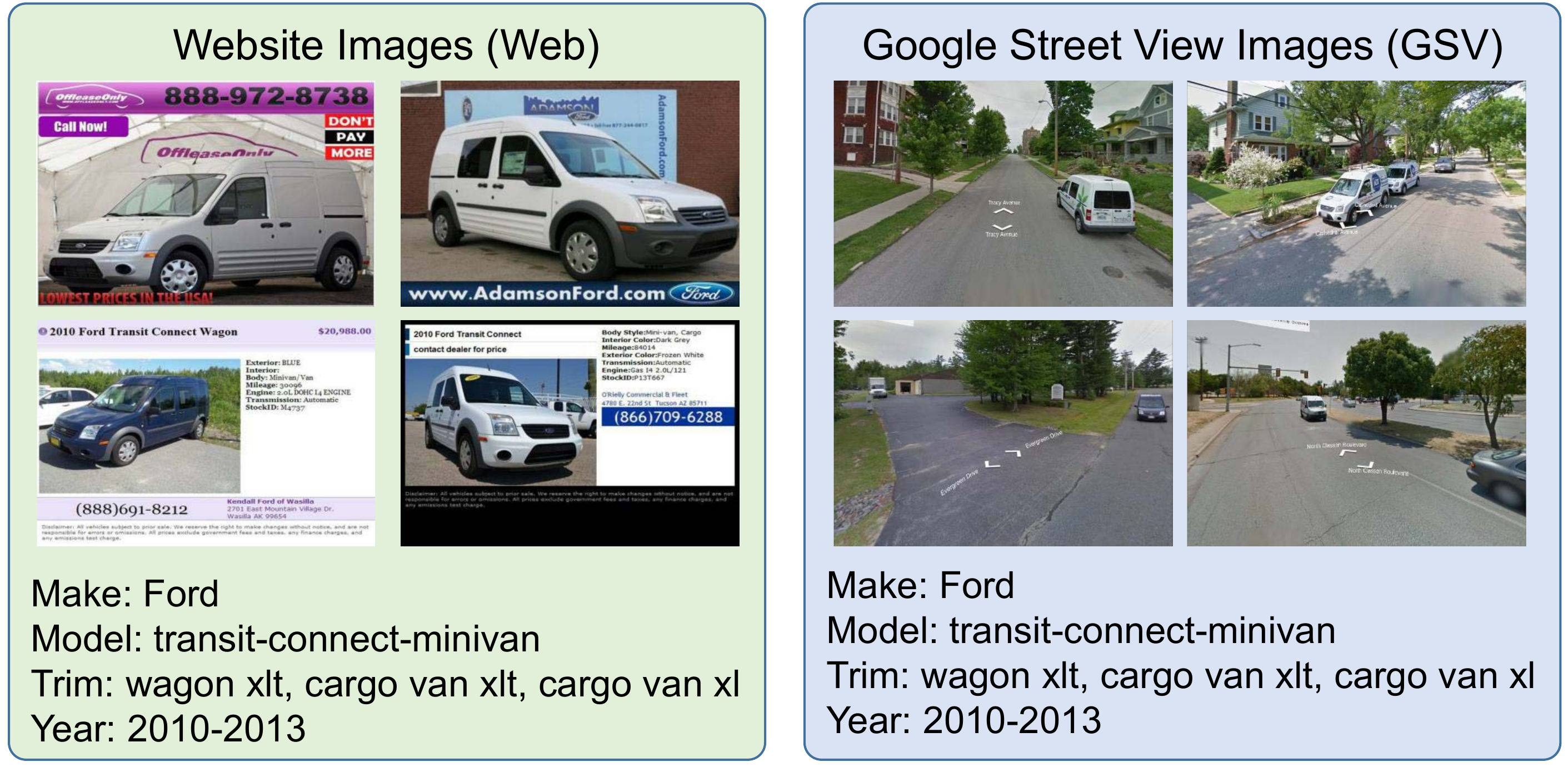}
\caption{ Examples of Website (Web) and Google Street View (GSV) Images for one type of car in the used fine-grained car dataset.}
\label{Fig_fine_grain_data}\vspace{-2mm}
\end{figure}
\begin{table*}[tb]\footnotesize
  \centering
  \caption{Accuracy (\%) of DRCN, DRCN ($\alpha=0$), DRCN ($\beta=0$), DRCN ($\w=\mathbf{1}$) and DRCN (w/o RCB) on Office-Home for \textbf{partial} transfer learning tasks (ResNet-50).}
   \setlength{\tabcolsep}{1.0mm}{
    \begin{tabular}{|c|ccccccccccccc|}
    \hline
    \multirow{2}[4]{*}{Method} & \multicolumn{13}{c|}{Office-Home} \bigstrut\\
\cline{2-14}          & Ar$\rightarrow$Cl  & Ar$\rightarrow$Pr  & Ar$\rightarrow$Rw  & Cl$\rightarrow$Ar  & Cl$\rightarrow$Pr  & Cl$\rightarrow$Rw  & Pr$\rightarrow$Ar  & Pr$\rightarrow$Cl  & Pr$\rightarrow$Rw  & Rw$\rightarrow$Ar  & Rw$\rightarrow$Cl  & Rw$\rightarrow$Pr  & Average \bigstrut\\
    \hline
    IWAN  & 53.9  & 54.5  & 78.1  & 61.3  & 48.0  & 63.3  & 54.2  & \textbf{52.0}  & \textbf{81.3} & 76.5  & 56.8  & \textbf{82.9} & 63.6  \bigstrut[t]\\
    SAN   & 44.4  & 68.7  & 74.6  & \textbf{67.5} & 65.0  & \textbf{77.8} & 59.8  & 44.7  & 80.1  & 72.2  & 50.2  & 78.7  & 65.3  \\
    PADA  & 52.0  & 67.0  & 78.7  & 52.2  & 53.8  & 59.0  & 52.6  & 43.2  & 78.8  & 73.7  & 56.6  & 77.1  & 62.1  \bigstrut[b]\\
    \hline
    \textbf{DRCN($\alpha=0$)} & 46.0  & 65.8  & 81.0  & 49.0  & 51.2  & 60.2  & 59.1  & 33.8  & 78.4  & 71.4  & 45.8  & 76.5  & 59.9  \bigstrut[t]\\
    \textbf{DRCN($\beta=0$)} & 49.0  & 68.4  & 79.5  & 56.1  & 56.0  & 64.3  & 61.0  & 43.5  & 74.3  & 66.4  & 50.7  & 73.2  & 61.9  \\
    \textbf{DRCN($\w=\mathbf{1}$)} & \textbf{54.2} & 69.7  & 82.7 & 51.5  & 65.7 & 62.4  & 61.3  & 44.3  & 76.8  & 77.8 & 50.9  & 76.1  & 64.4  \\
    \textbf{DRCN (w/o RCB)} & 53.2  & 73.3  & 80.9  & 59.5  & 58.7  & 68.0  & 64.6  & 48.7  & 78.7  & 73.6  & 56.6  & 79.4  & 66.3  \\
    \textbf{DRCN (soft label)} & \textbf{54.0} & \textbf{76.4} & \textbf{83.1}  & 62.1  & \textbf{66.3}  & 71.0  & \textbf{68.3} & 49.8  & 80.4  & \textbf{77.9}  & \textbf{58.1}  & 79.9  & \textbf{68.9} \\
    \textbf{DRCN (hard label)} & 51.6  & 75.8  & 82.0 & 62.9  & 64.1 & 72.9  & 67.4  & 50.0  & 81.0  & 76.4 & 57.7  & 79.3  & 68.4 \bigstrut[b]\\
    \hline
    \end{tabular}}%
  \label{tab:variants_of_drcn_officehome}
\end{table*}%
\begin{figure}[!t]
\centering
\includegraphics[width=0.5\textwidth]{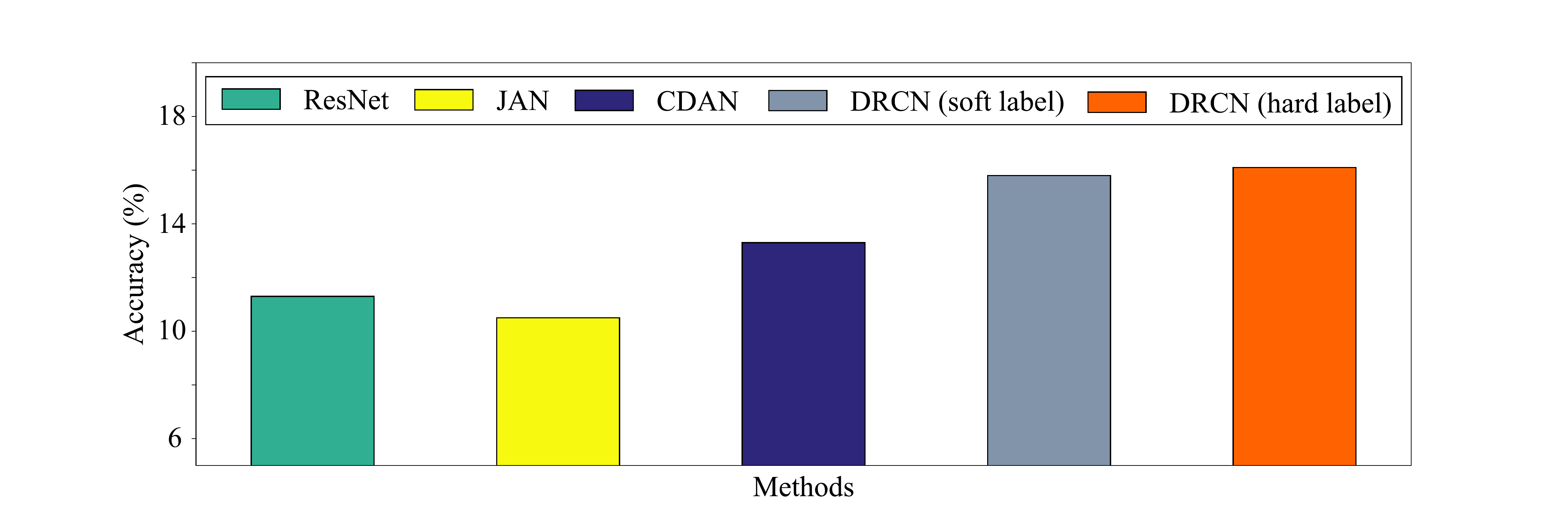}
\caption{ Classification accuracies of ResNet, JAN, CDAN and DRCN on the task \textbf{Web} $\rightarrow$ \textbf{GSV} of fine-grained recognition in the wild.}
\label{Fig_fine_grain_result}\vspace{-4mm}
\end{figure}
\subsubsection{Cross-domain Fine-grained Recognition in the Wild}\label{sec:fine-grained}

Fine-grained visual recognition is a well-studied problem to distinguish between objects in the same category (e.g. different car brands) \cite{fine-grained1,fine-grained2,fine-grained-TL,Fine-grained-car,MTDA}. However, it is infeasible to annotate enough data for every new scenario, thus addressing the problem of cross-domain fine-grained object recognition in the wild is more challenging and practical in computer vision.
%
% Table generated by Excel2LaTeX from sheet 'b=0,w=1,partial'

%
Similar to \cite{MTDA}, we can adapt the discriminative knowledge of annotated sources such as e-commerce websites to a sparse set of annotations in the real world, i.e. Google Street View. Fig. \ref{Fig_fine_grain_data} illustrates the examples of a fine-grained car dataset introduced in \cite{Fine-grained-car,MTDA}, which totally contains 1,095,021 images with more than 2,000 classes. The car images from craigslist.com, cars.com and edmunds.com are referred to as \textbf{Web} images, which are with high resolution and typically un-occluded. By contrast, the car images from Google Street View are blurry and occluded, which are denoted as \textbf{GSV} images.

It is apparently to observe that, Web and GSV images vary a lot in viewpoint, occlusion and resolution, leading to a large distribution discrepancy between these two sets. To evaluate the effectiveness of DRCN on the problem of fine-grained recognition in the wild, we construct the adaptation task \textbf{Web} $\rightarrow$ \textbf{GSV}. Here, we also use ResNet-50 pre-trained on ImageNet as the basic network. The results are illustrated in Fig. \ref{Fig_fine_grain_result}.
From the results, we can clearly obtain that DRCN outperforms other comparisons on this challenging fine-grained car recognition in the wild dataset, and gain $\mathbf{42.5\%}$ and $\mathbf{21.1\%}$ relative increases when compared to ResNet \cite{resnet} and CDAN \cite{CDAN}, with respect to the target prediction accuracy. This manifests DRCN can be well generalized to the problems of fine-grained car categorization in the wild, and significantly boosts the universality of DRCN.

\subsection{Empirical Analysis}\label{sec:Analytical-Experiments}
In this section, we will conduct several empirical experiments to verify the effectiveness of DRCN in detail.
\subsubsection{Feature Visualization}

\begin{figure*}[!t]
\centering
\includegraphics[width=\textwidth]{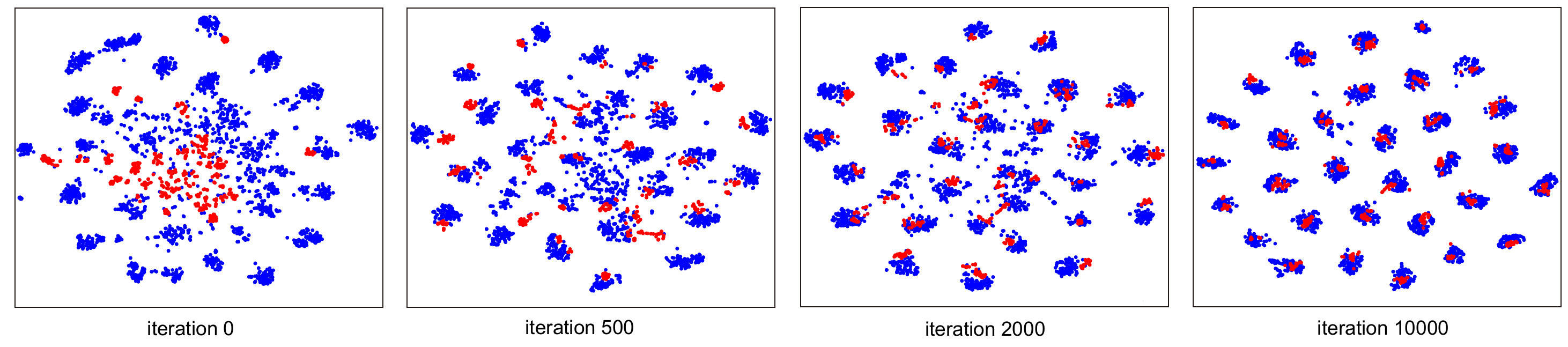}\vspace{-4mm}
\caption{t-SNE visualization of source and target corrected features for \textbf{traditional} domain adaptation in task A $\rightarrow$ W (31 classes) at different training iterations. Source data are blue dots and red dots represent target data.}
\label{Fig4_tsne_traditional}\vspace{-3mm}
\end{figure*}

\begin{figure*}[!t]
\centering
\includegraphics[width=\textwidth]{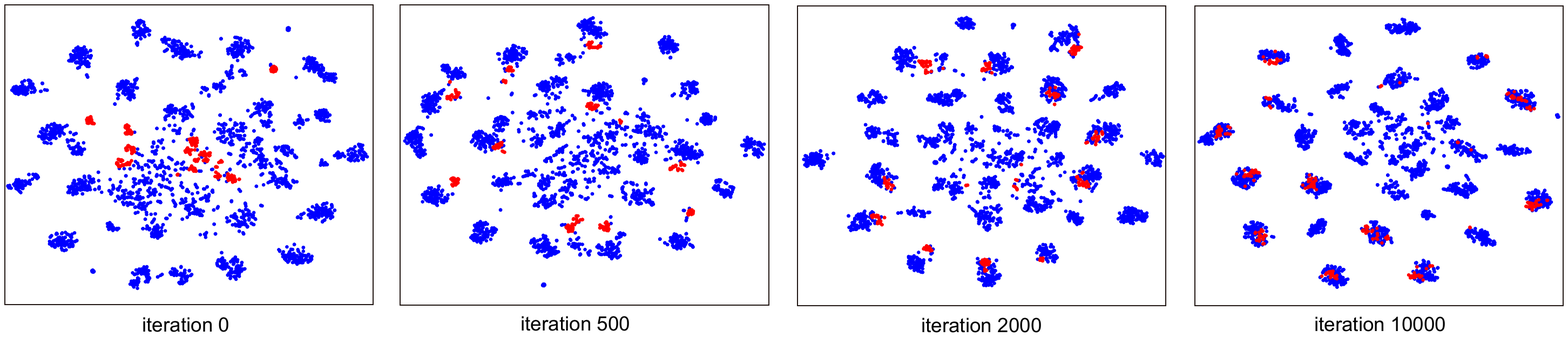}\vspace{-4mm}
\caption{t-SNE visualization of source and target corrected features for \textbf{partial} domain adaptation in task A $\rightarrow$ W (31 classes) at different training iterations. Source data are blue dots and red dots represent target data.}
\label{Fig5_tsne_partial}\vspace{-3mm}
\end{figure*}

\begin{figure*}[!t]
\centering
\includegraphics[width=\textwidth]{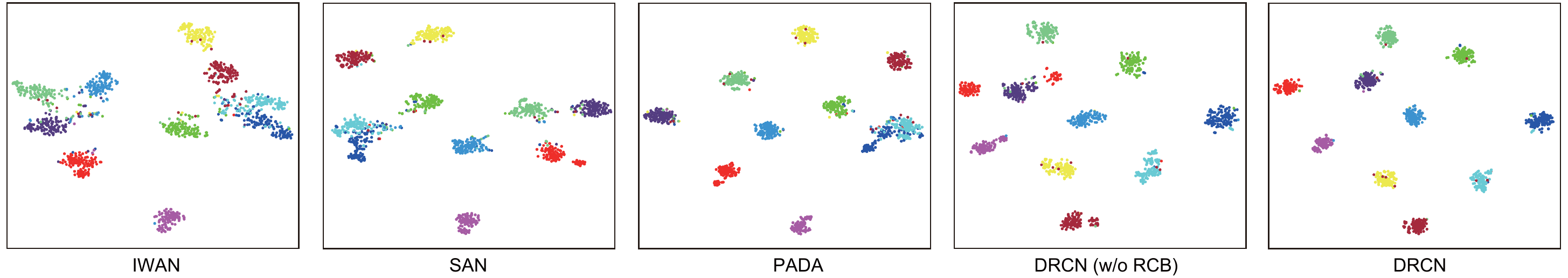}\vspace{-4mm}
\caption{t-SNE visualization of different methods (IWAN \cite{WAN-for-partial}, SAN \cite{SAN}, PADA \cite{PADA}, DRCN w/o RCB, DRCN) for \textbf{partial} domain adaptation of task W $\rightarrow$ A (10 classes) in the shared class space. Each class is represented with different color.}\vspace{-3mm}
\label{Fig6_tsne_parital_methods_comparison}
\end{figure*}

To effectively present the feature correction process of DRCN, in Fig. \ref{Fig4_tsne_traditional} and Fig. \ref{Fig5_tsne_partial}, we visualize the t-SNE embeddings \cite{tsne} of source and target corrected representations for \textbf{traditional} and \textbf{partial} domain adaptation scenarios (task: A $\rightarrow$ W in Office-31) at different learning iterations. From Fig. \ref{Fig4_tsne_traditional} and Fig. \ref{Fig5_tsne_partial}, we have the following observations. First, in the beginning, the source data are discriminated well while target data are substantially different with source, which demonstrates that there exists a large domain shift between source and target. Second, as the training process continues, the source and target representations become similar and each class across domains begins to align gradually. We attribute this phenomenon to the domain-wise and class-wise knowledge transfer and the powerful learning capacity of the plugged correction residual block. Third, by the end of training, source and target samples are nearly indistinguishable for both traditional and partial transfer scenarios, which manifests the effectiveness of DRCN to deal with domain adaptation problems.

Fig. \ref{Fig6_tsne_parital_methods_comparison} presents the t-SNE \cite{tsne} visualization of the task-specific features learned by baselines, and of the features before and after the residual correction block (RCB) for DRCN in the shared class space. In IWAN and SAN, the data structures are more scattered, as there still has many outlier samples cannot converge into their class clusters. PADA can successfully mitigate domain shift. However, some categories are mixed. Different from them, DRCN can not only successfully identify the most relevant classes but also represent good cohesion in clusters. These results verify the efficacy and superiority of DRCN comparing to other methods.
Note that the feature learned by DRCN w/o RCB is not quite compact and some class centers are misaligned. However, they become more clustered after passing RCB in DRCN. This improvement benefits from the insertion of RCB, which further mitigates feature discrepancy.

\subsubsection{Variants of DRCN}

\begin{table}[htbp]\scriptsize
  \centering
  \caption{Accuracy (\%) of DRCN, DRCN ($\alpha=0$), DRCN ($\beta=0$), DRCN ($\w=\mathbf{1}$) and DRCN (w/o RCB) on Office-31 for \textbf{partial} transfer learning tasks (ResNet-50).}\vspace{-3mm}
  \setlength{\tabcolsep}{1.0mm}{
    \begin{tabular}{|c|ccccccc|}
    \hline
    \multirow{2}[4]{*}{Method} & \multicolumn{7}{c|}{Office-31} \bigstrut\\
\cline{2-8}          & A$\rightarrow$W    & D$\rightarrow$W    & W$\rightarrow$D    & A$\rightarrow$D    & D$\rightarrow$A    & W$\rightarrow$A    & Average \bigstrut\\
    \hline
    IWAN  & 89.2  & 99.3  & 99.4  & 90.5  & \textbf{95.6} & 84.7  & 93.1  \bigstrut[t]\\
    SAN   & \textbf{93.9} & 99.3  & 99.4  & 94.3  & 94.2  & 88.7  & 95.0  \\
    PADA  & 86.5  & 99.3  & \textbf{100.0} & 82.2  & 92.7  & 95.4  & 92.7  \bigstrut[b]\\
    \hline
    \textbf{DRCN($\alpha=0$)} & 89.0  & 99.3  & \textbf{100.0} & 93.6  & 89.1  & 93.3  & 94.1  \bigstrut[t]\\
    \textbf{DRCN($\beta=0$)} & 86.4  & \textbf{100.0} & \textbf{100.0} & 84.7  & 93.4  & 92.5  & 93.3 \\
    \textbf{DRCN($\w=\mathbf{1}$)} & 86.7  & 98.3  & \textbf{100.0} & 86.0  & 94.1  & 94.4  & 93.2  \\
    \textbf{DRCN (w/o RCB)} & 86.8  & \textbf{100.0} & 99.6  & 86.6  & 93.4  & 93.4  & 93.3  \\
    \textbf{DRCN (soft label)} & 88.5  & \textbf{100.0} & \textbf{100.0} & 86.0  & \textbf{95.6} & \textbf{95.8} & 94.3  \\
    \textbf{DRCN (hard label)} & 90.8  & \textbf{100.0} & \textbf{100.0} & \textbf{94.3} & 95.2  & 94.8  & \textbf{95.9} \bigstrut[b]\\
    \hline
    \end{tabular}}%
  \label{tab:variants_of_drcn_office31}%
\end{table}%

We carry out four variants of DRCN (soft label) to go deeper into the influence of parameter $\alpha$ and $\beta$, the component residual correction block, as well as weights $\w$ on the classification performance. The results are reported in Tables \ref{tab:variants_of_drcn_officehome} and \ref{tab:variants_of_drcn_office31}. To be specific, DRCN ($\alpha=0$) is the variant without loss of domain-wise knowledge transfer, i.e. without $\mathcal{L}_{domain}$. Similarly, DRCN ($\beta=0$) is the variant without loss of class-wise knowledge transfer, while DRCN (w/o RCB) does not have residual correction block. DRCN ($\w=\mathbf{1}$) is the variant in which its each class weight $\w^{(k)}$ is 1 in Eq (\ref{eq11:weighted-class-mmd}) with equal importance contribution for each class in terms of the class-wise alignment loss $\mathcal{L}_{class}$. Additionally, we test the performances of DRCN and its variants under \textbf{partial} transfer tasks on Office-Home and Office-31 datasets.

The results in Tables \ref{tab:variants_of_drcn_officehome} and \ref{tab:variants_of_drcn_office31} reveal the following observations. First, it is obvious that DRCN (hard label), DRCN (soft label) and other variants of DRCN, except DRCN ($\alpha=0$), all outperform PADA. This fact indicates that minimizing the domain-wise distribution difference is also important for tackling partial domain adaptation problems.
Second, by comparing DRCN with DRCN ($\beta=0$), we observe that DRCN outperforms DRCN ($\beta=0$) with over 1\% improvement in Office-31, even achieving 7 \% in Office-Home, demonstrating that our designed weighted class-wise alignment is an essential part in DRCN.
%and DRCN particularly attains the greatest improvement of 2.1\%.
Third, we can find out that DRCN (hard label) and DRCN (soft label) overpass IWAN and SAN with a large margin, whereas DRCN ($\w=\mathbf{1}$) is worse than SAN.
This is reasonable if we fix all class weight as 1, which would easily trigger negative transfer from irrelevant classes.
%Besides, when classifier can not precisely produce class probabilities, using soft label will lead to unexpected class confusion, which can explain DRCN ($\w=\mathbf{1}$) is slightly better than DRCN on few tasks.
%
Fourth, DRCN outperforms DRCN (w/o RCB) in both Office-31 and Office-Home. In Office-31, we can see over 1\% accuracy decrease in DRCN (w/o RCB). The performance decline is more serious in Office-Home, with DRCN (w/o RCB) degrading more than 2\%. Thus, the effectiveness of Residual Correction Block (RCB) can be successfully verified.

As a result, we can safely conclude that both class-wise knowledge and domain-wise knowledge need to be transferred simultaneously. Furthermore, residual correction block indeed helps enhance DRCN with greater capacity to knowledge transfer, and weights assigned to class is crucial to enabling positive transfer in the shared label space. Therefore, each component in DRCN is trying to solve partial domain adaptation from distinct aspect.
%Therefore to avoid this problem, class-wise weights can help us to align the relevant classes effectively.
%
%Similar to partial transfer tasks, DRCN ($\alpha=0$) takes the lowest accuracy among compared methods on traditional transfer tasks due to the lack of valuable general domain-wise feature alignment.
%By adding proposed weighted class-wise alignment with soft label, DRCN can meet performance which is comparable to other advanced methods.
%Those results prove that DRCN is versatile to both \textbf{partial} and \textbf{traditional} transfer scenarios.
%because the across-domain distribution is more different on this dataset and the soft label given by the classifier at this time cannot help us to align the relevant classes effectively. Therefore, it is important to transfer general knowledge to the target domain which can be achieved by fixing all class weight as 1.
%However, on the easier dataset Office-31 class-wise alignment has been done well, properly adjusting the weight of both, we are likely to get better results than DRCN, which is why that DRCN is worse than DRCN ($\alpha=0$) on the task A$\rightarrow$W and A$\rightarrow$D.

%weighted class-wise distribution matching to improve the positive effects of relevant source classes and reduce the negative impacts of irrelevant source classes, which could benefit valuable knowledge transfer from source to target a lot.

\begin{figure}[!t]
\centering
\includegraphics[width=0.47\textwidth]{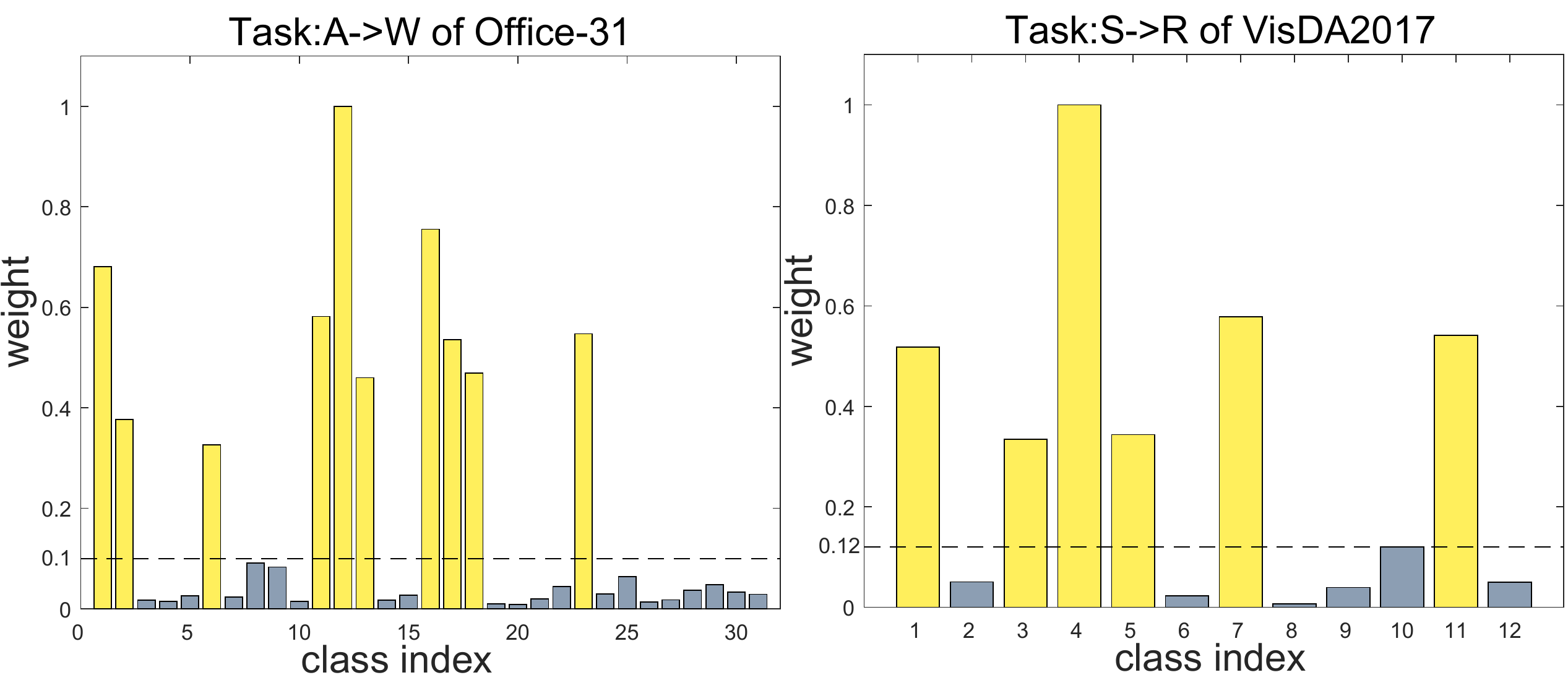}\vspace{-3mm}
\caption{Histograms of different class weights learned by DRCN for task: A $\rightarrow$ W of Office-31 and S $\rightarrow$ R of VisDA2017.}\vspace{-5mm}
\label{Fig7_weight}
\end{figure}

\subsubsection{Statistics of Class Weights}

For partial domain adaptation, to verify the effectiveness of the weighted scheme in DRCN, Fig. \ref{Fig7_weight} presents the learned class weights for task: A $\rightarrow$ W of Office-31 and S $\rightarrow$ R of VisDA2017. The yellow bars represent the weights of shared classes, and grey bars are for irrelevant classes.

It is inspiring to see that the weights of shared classes are much larger than that of irrelevant classes. For task A $\rightarrow$ W as an example, all the weights of irrelevant classes are less than 0.1, and even the smallest weight of shared class is still several times of the largest weight of irrelevant classes. This weighted strategy by leveraging the target output distribution indeed can identify the most similar source subclasses to target domain, and improve the positive effects of them, simultaneously reduce the influence of irrelevant source data. Then, the weighted class-wise matching in DRCN will transfer the most relevant knowledge from a large-scale source domain to a small-scale target domain effectively.

\subsubsection{Parameter Sensitivity}

\begin{figure}[tb]
\centering
\includegraphics[width=0.5\textwidth]{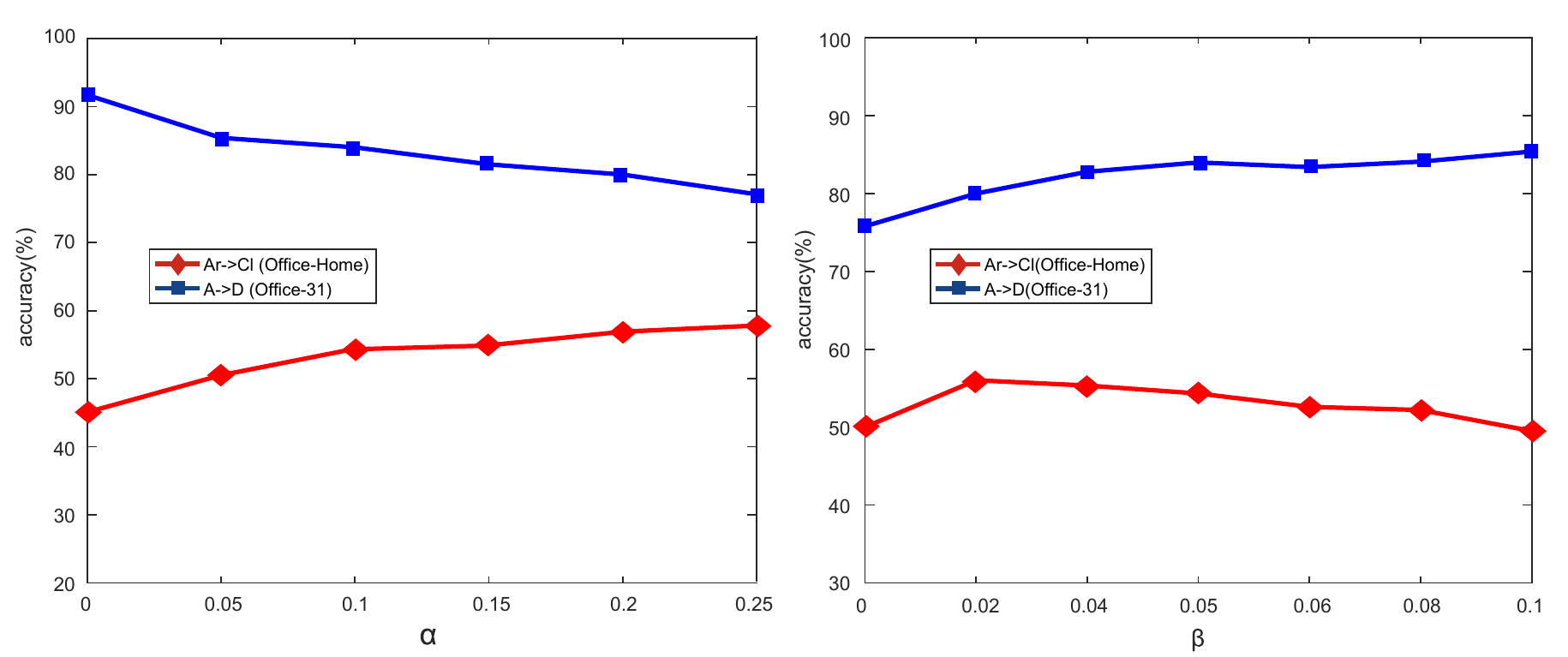}\vspace{-3mm}
\caption{Parameter sensitivity studies w.r.t. $\alpha$ and $\beta$ respectively. }\vspace{-3mm}
\label{Fig9_parameter}
\end{figure}

As shown in Fig. \ref{Fig9_parameter}, we want to investigate the effects of parameters $\alpha$ and $\beta$ on the experimental results by varying $\alpha \in \{0, 0.05, 0.1, 0.15, 0.2, 0.25\}$ and $\beta \in \{0, 0.02, 0.04, 0.05, 0.06, 0.08, 0.1\}$. Besides, we use two random transfer tasks which are \textbf{Ar$\rightarrow$Cl} (Office-Home) and \textbf{A$\rightarrow$D} (Office-31) to testify the performance.

Concerning with the parameter $\alpha$, we observe that as $\alpha$ varies from 0 to 0.25, the classification accuracy of Office-31 decreases, while that of Office-Home increases. It is desirable that when Office-31 dataset has been aligned with the source and target domains, increasing $\alpha$, in other words, preserving more general knowledge, will lead to negative transfer. However, the domains in Office-Home are very dissimilar with each other, only transferring task-level knowledge would not enough unless taking general knowledge into consideration. Therefore, for easy transfer tasks, we can safely choose a small $\alpha$, but for hard transfer tasks, a relative large $\alpha$ will facilitate the general domain knowledge transfer.

As for the parameter $\beta$, the accuracy in Office-31 shows a growing trend with the increase of $\beta$, which further validates the importance of strengthening class-wise alignment under circumstances similar to Office-31. While the accuracy of Office-Home increases first and then decreases slightly as $\beta$ increases. Note that there is a slight increase when $\beta = 0.02$, but the overall trend is still decreasing. An interpretation is that the average accuracy of Office-Home on the task \textbf{Ar$\rightarrow$Cl} is very low, which indicates most of class-wise alignment can be incorrect. Therefore, increasing $\beta$ means emphasizing the contribution of misalignment, and that will lead to more false classification results. Those observations demonstrate that proper trade-off will enhance effective knowledge transfer in DRCN.

Thus, if we choose reasonable values for $\alpha$ and $\beta$, DRCN could perform stably and outperform other competitors.
% bell-shaped curve
%this suggests that  proper trade-off enhance transferability.
% balance the contribution between loss of domain and class-wise weight

\subsubsection{Layer Responses Analysis}

%From Fig. \ref{Fig4_tsne_traditional} to Fig. \ref{Fig6_tsne_parital_methods_comparison}, we observe that the added residual correction block indeed mitigates the discrepancy between source and target.
As shown in Fig. \ref{Fig8_mean_var_adistance} (a), to quantitatively characterize how much information the added residual block has learned, we calculate the mean and variance of the task-specific layer response $F_s(\x_t)$ and $F_s(\x_s)$, added residual correction layer response $\Delta F_s(\x_t)$ and their element-wise summation $F_t(\x_t)$ on task: W $\rightarrow$ A (Office-31), respectively.

The results show that the added residual block indeed has learned the discrepancy between source and target. Because the mean and variance values of $F_t(\x_t)$, which are calculated using features after residual correction block, are much similar with that of $F_s(\x_s)$, whereas there exists a relatively large gap between $F_s(\x_t)$ and $F_s(\x_s)$ in their values. Therefore, if the target data only passes through the original source network, the domain shift couldn't be mitigated so well without residual correction. This result manifests the added residual correction block has the powerful learning capacity for effective adaptation.

\subsubsection{Proxy $\mathcal{A}$-distance for Traditional Domain Adaptation}

\begin{figure}[!t]
\centering
\includegraphics[width=0.5\textwidth]{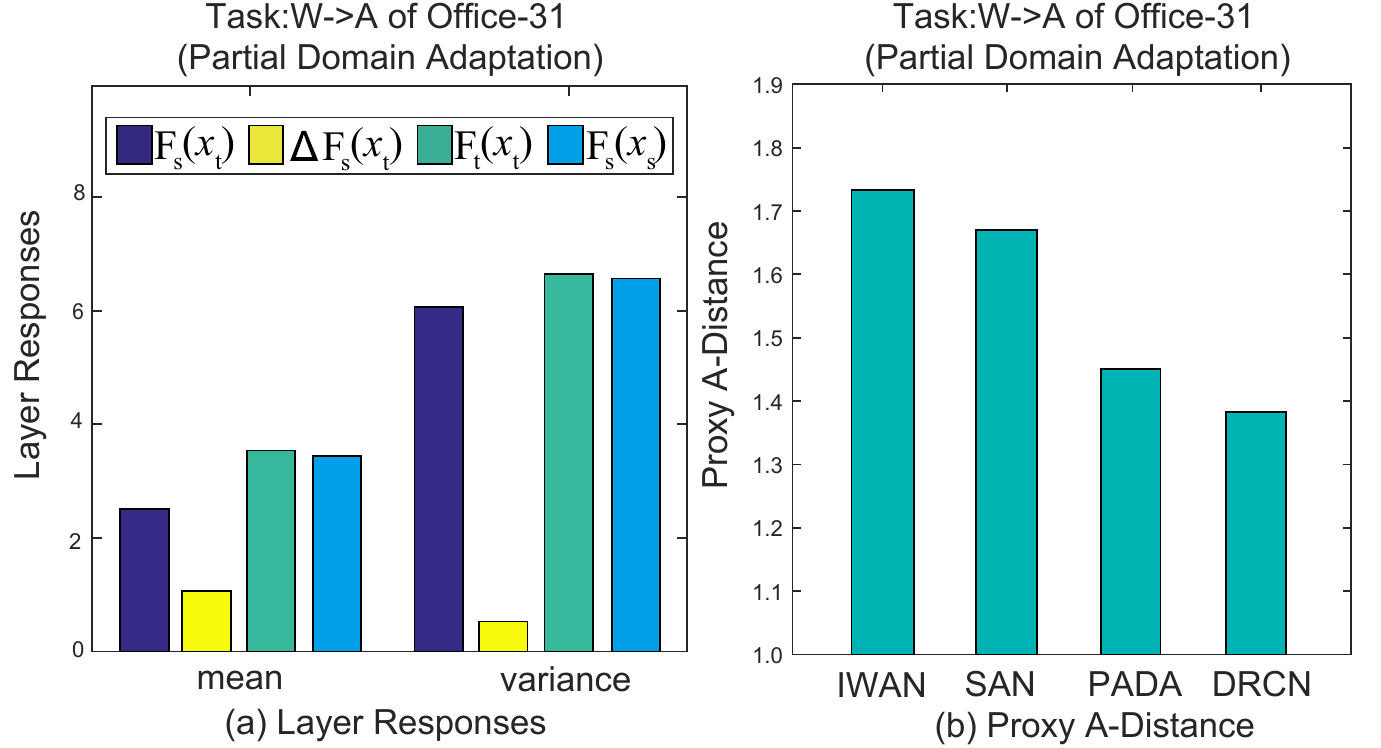}\vspace{-3mm}
\caption{(a) Statistics of task-specific layer response $F_s(\x_t)$ and $F_s(\x_s)$ , added residual correction layer response $\Delta F_s(\x_t)$ and their element-wise summation $F_t(\x_t)$ on task: W $\rightarrow$ A of Office-31 (\textbf{partial} domain adaptation); (b) Proxy $\mathcal{A}$-distance of IWAN \cite{WAN-for-partial}, SAN \cite{SAN}, PADA \cite{PADA} and DRCN features on task: W $\rightarrow$ A of Office-31 (\textbf{partial} domain adaptation).}\vspace{-3mm}
\label{Fig8_mean_var_adistance}
\end{figure}

$\mathcal{A}$-distance is a measure of distance between two distributions \cite{A-distance}, and a larger $\mathcal{A}$-distance implies it is easier to discriminate source and target. Since directly computing $\mathcal{A}$-distance is intractable, we resort to leverage proxy $\mathcal{A}$-distance instead. The proxy $\mathcal{A}$-distance is defined as
\begin{align}
\label{eq13:a-distance}
\widehat{d}_\mathcal{A}=2(1-2\epsilon),
\end{align}
where $\epsilon$ is the classification error of classifying source and target data with task-specific representations.

To evaluate the effectiveness of learned features for different methods, we calculate the proxy $\mathcal{A}$-distance of IWAN \cite{WAN-for-partial}, SAN \cite{SAN}, PADA \cite{PADA} features extracted after the task-specific layer, and DRCN features $F_s(\x_s), F_t(\x_t)$ for task: W $\rightarrow$ A of Office-31 (partial domain adaptation scenario). The results are shown in Fig. \ref{Fig8_mean_var_adistance}(b).

We observe that DRCN has the smallest $\widehat{d}_\mathcal{A}$ among the four methods, followed by PADA, SAN, and IWAN. It is known that the smaller the A-distance is, the easier the feature is to transfer. The $\widehat{d}_\mathcal{A}$ of PADA features is less than IWAN and SAN, which implies PADA features could learn transferable representations to confuse classifier. However, the $\widehat{d}_\mathcal{A}$ using DRCN features is smaller than that of PADA. We therefore can argue that DRCN features are more indistinguishable and can close the domain shift effectively, which explains the better performance of DRCN than competitive methods.

\section{Conclusion}\label{sec:conclusion}

This paper introduces a novel Deep Residual Correction Network (DRCN) to address partial domain adaptation problems, in which the target label space is a subset of source label space. DRCN could learn the different significance of source classes automatically by leveraging the target output probability distribution. Based on the learned weights, we propose a weighted class-wise matching strategy to explicitly align target data with the most relevant source subclasses, and maximally mitigate the discrepancy across domains. Different from other partial domain adaptation architectures, DRCN also jointly transfers general feature-level and task-level knowledge from source to target, since we find that properly transferring general knowledge can benefit the final classification significantly as well. To boost the adaptation ability of structure, DRCN plugs one residual correction block into the general source network along with the task-specific feature layer, which is easily implemented and efficiently generalized to new designed deep networks. DRCN can be easily extended to deal with traditional and fine-grained cross-domain visual recognition tasks. Extensive experimental results on several standard cross-domain datasets have demonstrated that DRCN outperforms several competitive deep domain adaptation approaches in both partial and traditional domain adaptation scenarios by a large margin.

%
%\ifCLASSOPTIONcompsoc
%  % The Computer Society usually uses the plural form
%\else
%  % regular IEEE prefers the singular form
%\fi

\section*{Acknowledgment}
This work is supported in part by the National Natural Science Foundation of China under Grant No. 61902028, and in part by the National Key Research and Development Plan of China under Grant No. 2018YFB1003701 and 2018YFB1003700.

\ifCLASSOPTIONcaptionsoff
  \newpage
\fi

%\bibliography{reference}
%\bibliographystyle{IEEEtran}

\begin{IEEEbiography}[{\includegraphics[width=1in,height=1.25in,clip,keepaspectratio]{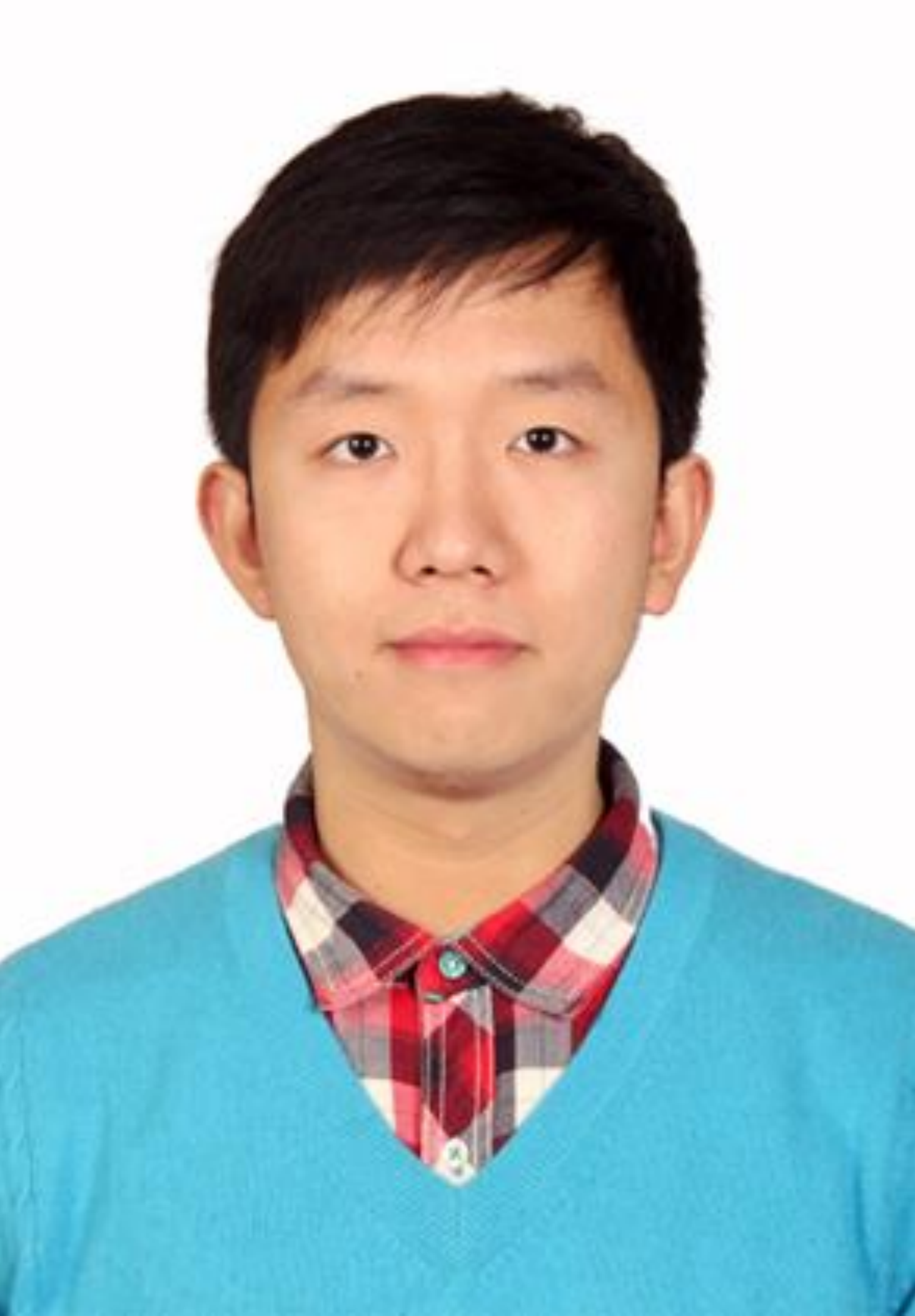}}]{Shuang Li} received the Ph.D. degree in control science and engineering from the Department of Automation, Tsinghua University, Beijing, China, in 2018.

He was a Visiting Research Scholar with the Department of Computer Science, Cornell University, Ithaca, NY, USA, from November 2015 to June 2016. He is currently an Assistant Professor with the school of Computer Science and Technology, Beijing Institute of Technology, Beijing. His main research interests include machine learning and deep learning, especially in transfer learning and domain adaptation.
\end{IEEEbiography}

\begin{IEEEbiography}[{\includegraphics[width=1in,height=1.25in,clip,keepaspectratio]{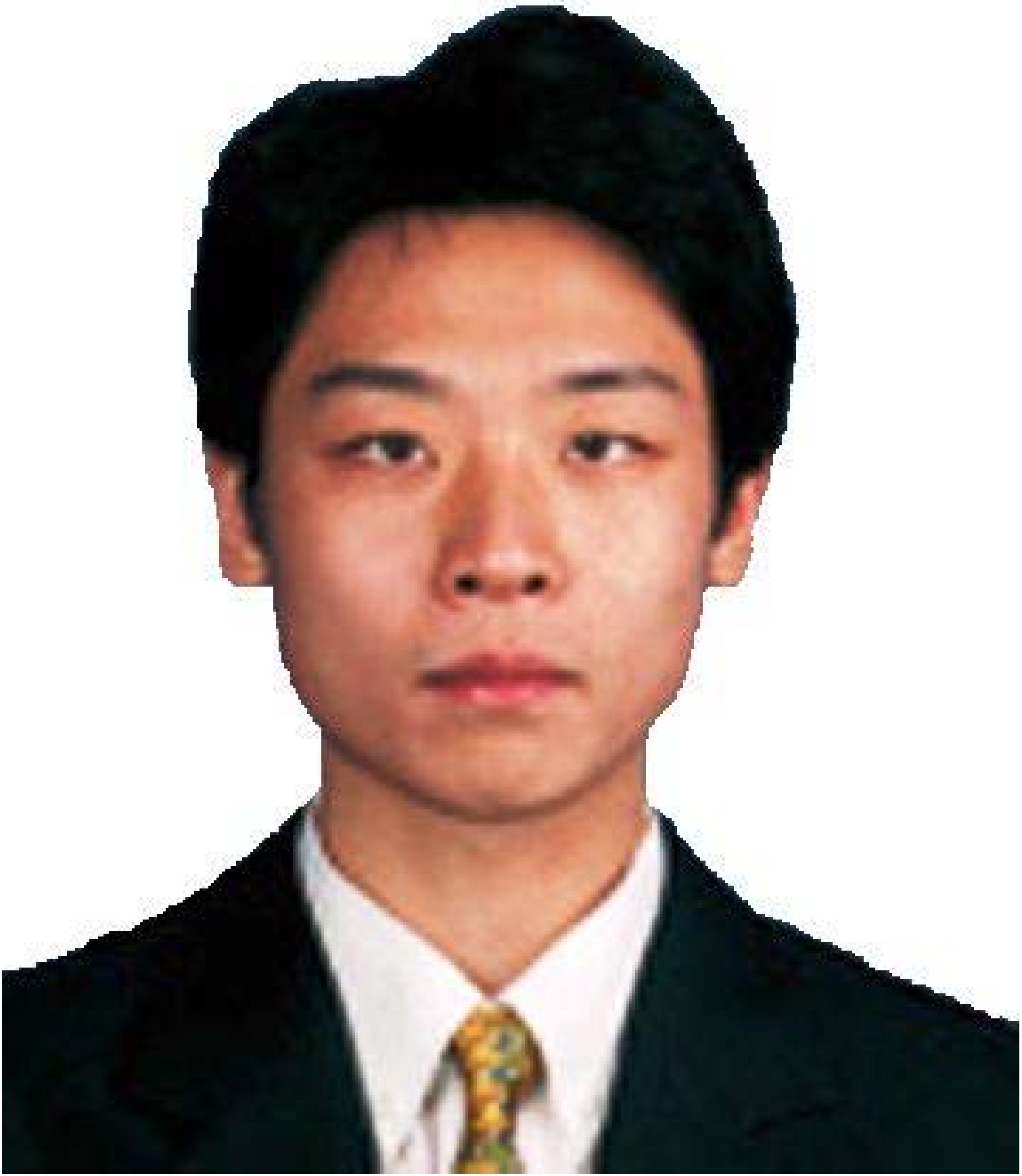}}]{Chi Harold Liu}(SM'15) receives the Ph.D. degree from Imperial College, UK in 2010, and the he B.Eng. de. degree from Tsinghua University, China in 2006. He is currently a Full Professor and Vice Dean at the School of Computer Science and Technology, Beijing Institute of Technology, China. He is also the Director of IBM Mainframe Excellence Center (Beijing), Director of IBM Big Data Technology Center, and Director of National Laboratory of Data Intelligence for China Light Industry. Before moving to academia, he joined IBM Research - China as a staff researcher and project manager, after working as a postdoctoral researcher at Deutsche Telekom Laboratories, Germany, and a visiting scholar at IBM T. J. Watson Research Center, USA. His current research interests include the Internet-of-Things (IoT), Big Data analytics, mobile computing, and deep learning. He received the Distinguished Young Scholar Award in 2013, IBM First Plateau Invention Achievement Award in 2012, and IBM First Patent Application Award in 2011 and was interviewed by EEWeb.com as the Featured Engineer in 2011. He has published more than 80 prestigious conference and journal papers and owned more than 14 EU/U.S./U.K./China patents. He serves as the Area Editor for KSII Trans. on Internet and Information Systems and the book editor for six books published by Taylor \& Francis Group, USA and China Machinery Press. He also has served as the general chair of IEEE SECON'13 workshop on IoT Networking and Control, IEEE WCNC'12 workshop on IoT Enabling Technologies, and ACM UbiComp'11 Workshop on Networking and Object Memories for IoT. He served as the consultant to Asian Development Bank, Bain \& Company, and KPMG, USA, and the peer reviewer for Qatar National Research Foundation, and National Science Foundation, China. He is a Senior Member of IEEE.
\end{IEEEbiography}

\begin{IEEEbiography}[{\includegraphics[width=1in,height=1.25in,clip,keepaspectratio]{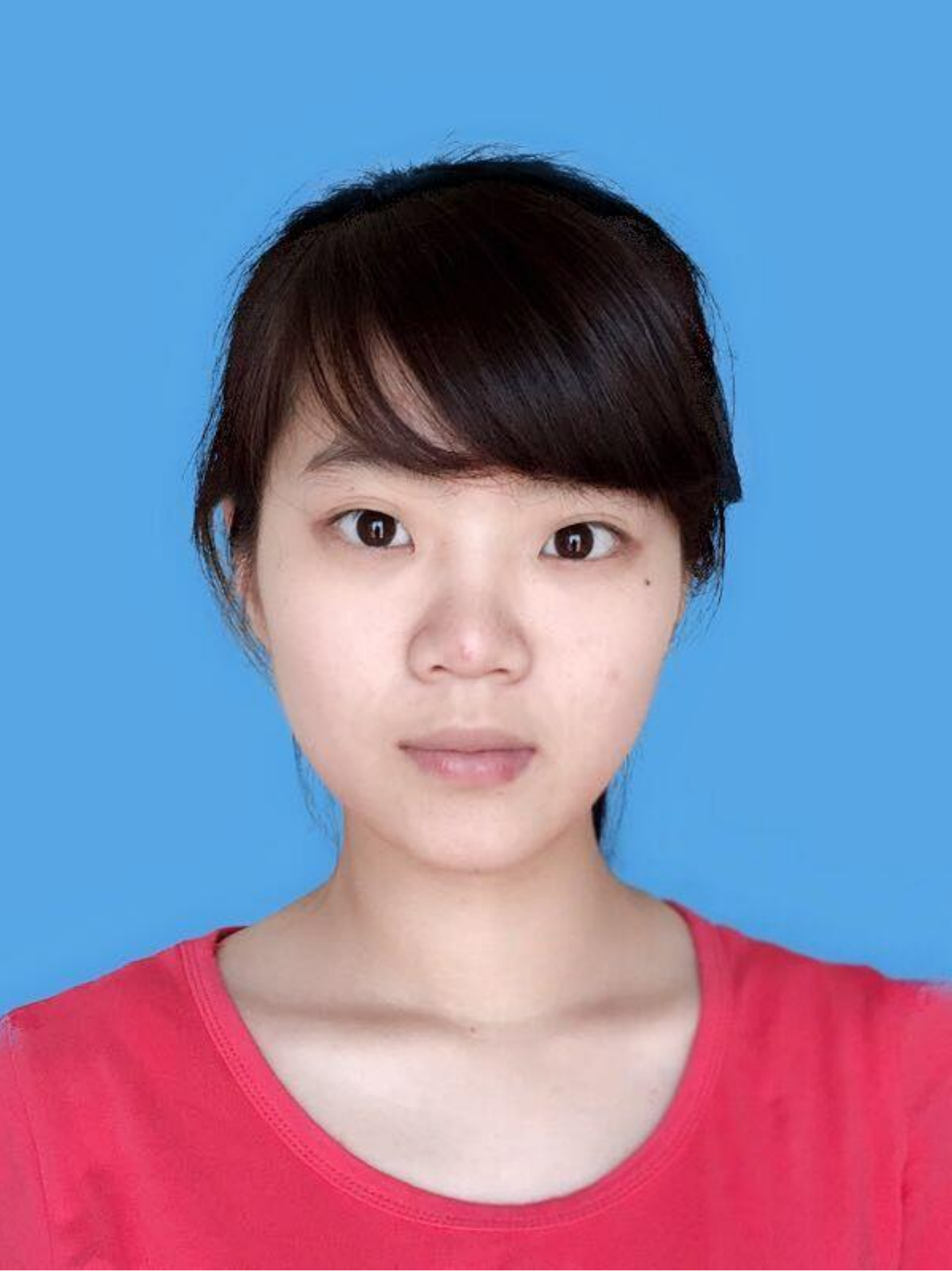}}]{Qiuxia Lin} is pursuing the M.S. degree in Computer Science from Beijing Institute of Technology. Her research interests include deep learning and transfer learning.
\end{IEEEbiography}

\begin{IEEEbiography}[{\includegraphics[width=1in,height=1.25in,clip,keepaspectratio]{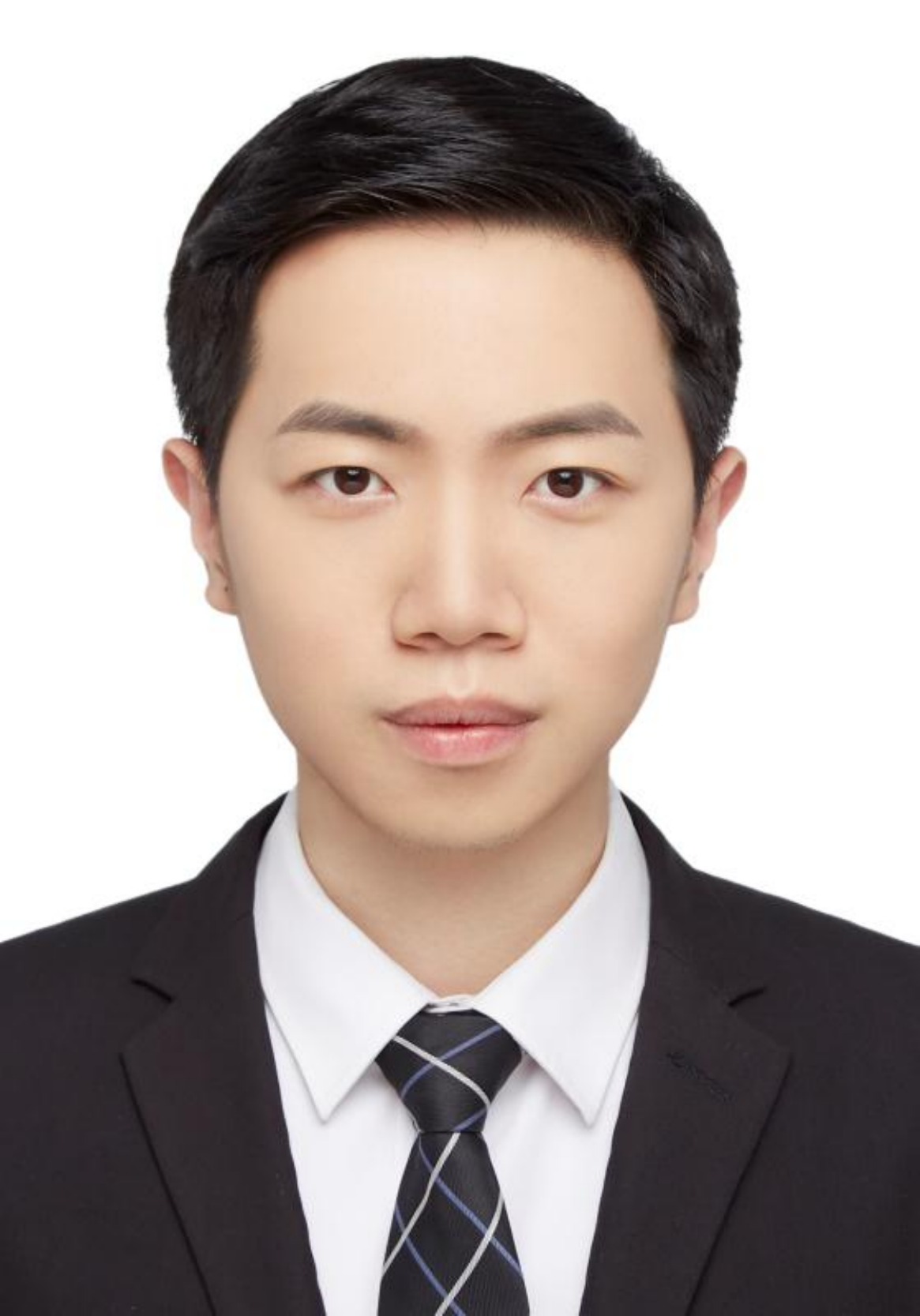}}]{Qi Wen} is pursuing the M.E. degree in Computer Science from Beijing Institute of Technology. His research interests include deep learning and transfer learning.
\end{IEEEbiography}

\begin{IEEEbiography}[{\includegraphics[width=1in,height=1.25in,clip,keepaspectratio]{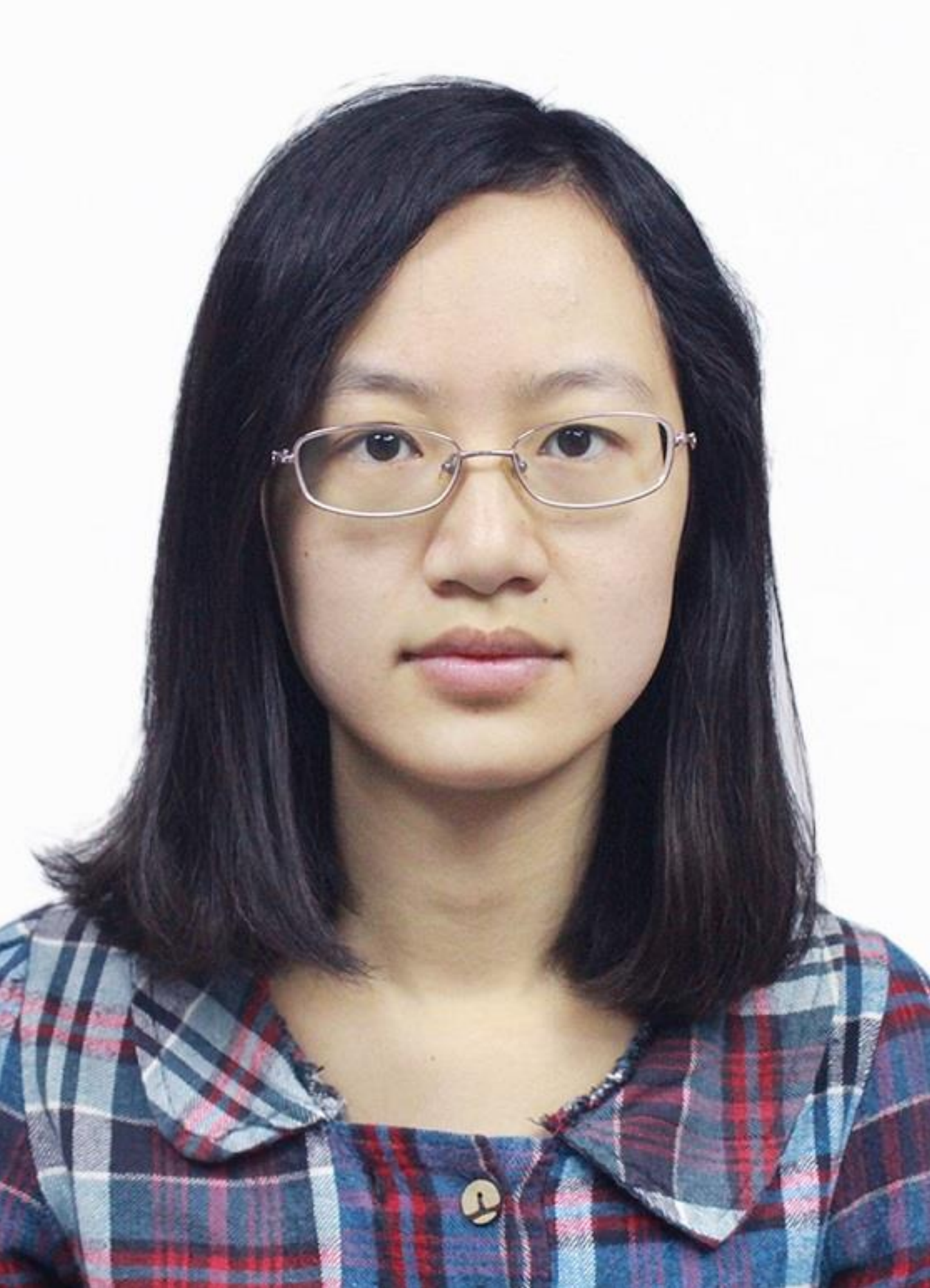}}]{Limin Su} is pursuing the B.E. degree in Computer Science from Beijing Institute of Technology. Her research interests include machine learning and transfer learning.
\end{IEEEbiography}

\begin{IEEEbiography}[{\includegraphics[width=1in,height=1.25in,clip,keepaspectratio]{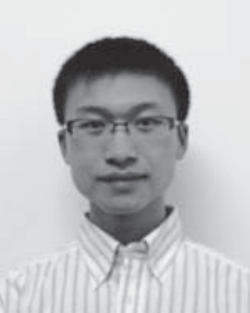}}]{Gao Huang} is an assistant professor in the Department of Automation, Tsinghua University. He was a Postdoctoral Researcher in the Department of Computer Science at Cornell University. He received the PhD degree in Control Science and Engineering from Tsinghua University in 2015, and B.Eng degree in Automation from Beihang University in 2009. He was a visiting student at Washington University at St. Louis and Nanyang Technological University in 2013 and 2014, respectively. His research interests include machine learning and computer vision.
\end{IEEEbiography}

\begin{IEEEbiography}[{\includegraphics[width=1in,height=1.25in,clip,keepaspectratio]{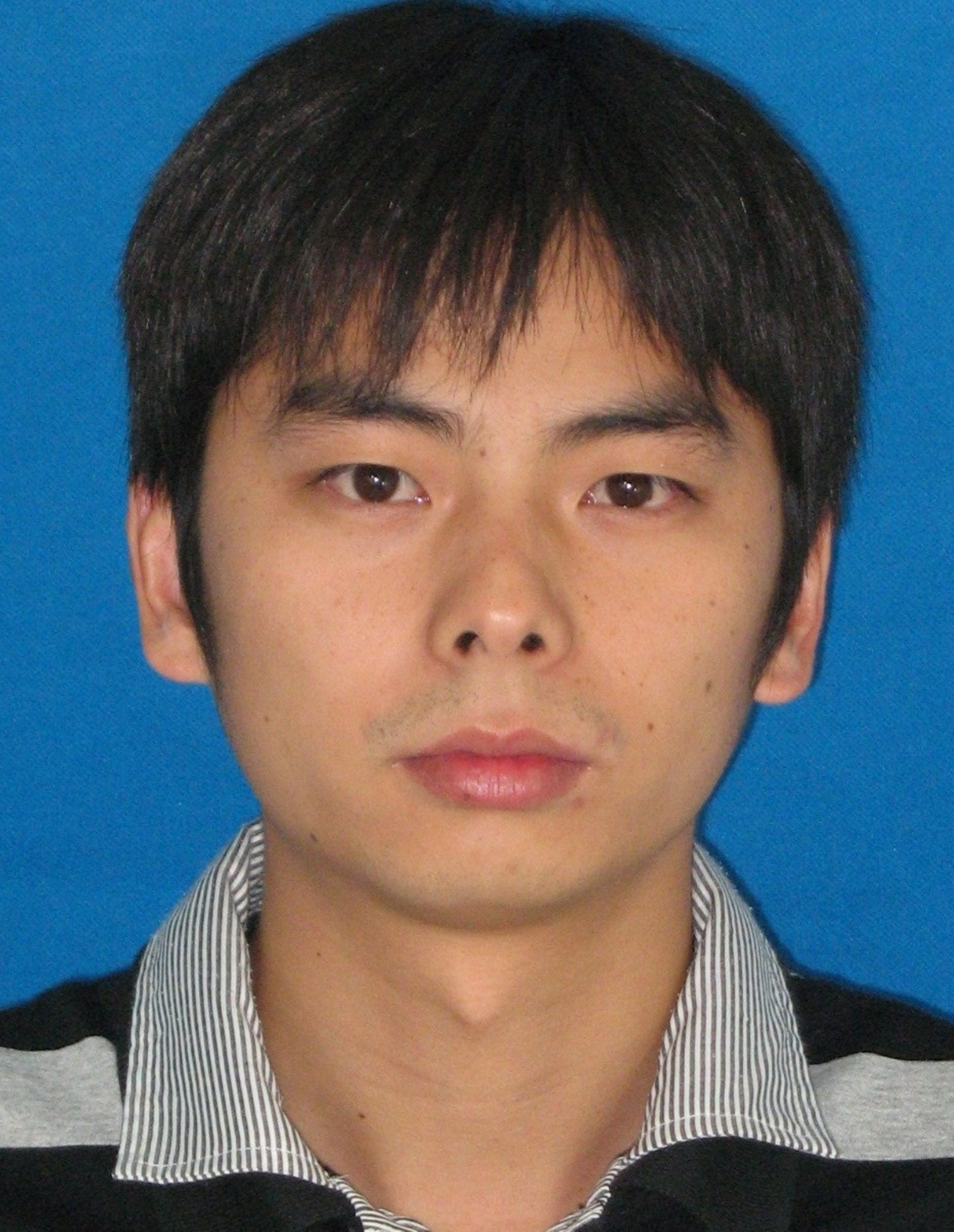}}]{Zhengming Ding}(S'14) received the B.Eng. degree in information security and the M.Eng. degree in computer software and theory from University of Electronic Science and Technology of China (UESTC), China, in 2010 and 2013, respectively. He received the Ph.D. degree from the Department of Electrical and Computer Engineering, Northeastern University, USA in 2018. He is a faculty member affiliated with Department of Computer, Information and Technology, Indiana University-Purdue University Indianapolis since 2018. His research interests include machine learning and computer vision. Specifically, he devotes himself to develop scalable algorithms for challenging problems in transfer learning and deep learning scenario. He received the National Institute of Justice Fellowship during 2016-2018. He was the recipients of the best paper award (SPIE 2016) and best paper candidate (ACM MM 2017). He is now an Associate Editor for the Journal of Electronic Imaging (JEI).
\end{IEEEbiography}

% \end{CJK}
\end{document}